\documentclass[sigconf, 10pt, nonacm=true, balance=false]{acmart}

\usepackage{listings}
\usepackage{color}
\usepackage{float}
\usepackage{wrapfig}
\usepackage{subcaption}

\definecolor{codegreen}{rgb}{0,0.6,0}
\definecolor{codegray}{rgb}{0,0,0}
\definecolor{codepurple}{rgb}{0.58,0,0.82}
\definecolor{backcolour}{rgb}{0.95,0.95,0.92}
\usepackage{xspace}
\usepackage{amsmath}

\usepackage{ifthen}
\usepackage{xcolor}
\newcommand{\exclude}[1]{}
\newcommand{\showComments}{yes}
\newcommand{\note}[2]{
    \ifthenelse{\equal{\showComments}{yes}}{\textcolor{#1}{#2}}{}
}

\newcommand{\sys}{\mbox{\textsc{MLDTW}}\xspace}
 
\lstdefinestyle{mystyle}{
    backgroundcolor=\color{backcolour},   
    commentstyle=\color{codegreen},
    keywordstyle=\color{magenta},
    numberstyle=\tiny\color{codegray},
    stringstyle=\color{codepurple},
    basicstyle=\fontsize{7}{9}\ttfamily,
    breakatwhitespace=false,         
    breaklines=true,                 
    captionpos=b,                    
    keepspaces=true,                 
    numbers=left,                    
    numbersep=5pt,                  
    showspaces=false,                
    showstringspaces=false,
    showtabs=false,                  
    tabsize=2,
    frame=single
}
 
\def\BibTeX{{\rm B\kern-.05em{\sc i\kern-.025em b}\kern-.08emT\kern-.1667em\lower.7ex\hbox{E}\kern-.125emX}}

\begin{document}

%
\title{Using Machine Learning to Augment Dynamic Time Warping Based Signal Classification}

%
\author{Arvind Seshan}
\affiliation{%
  \institution{}
  \city{Pittsburgh}
  \state{Pennsylvania}
}

%
\renewcommand{\shortauthors}{Seshan}

%

\begin{abstract}
Modern applications such as voice recognition rely on the ability to compare signals to pre-recorded ones to classify them. However, this comparison typically needs to ignore differences due to signal noise, temporal offset, signal magnitude, and other external factors. The Dynamic Time Warping (DTW) algorithm quantifies this similarity by finding corresponding regions between the signals and non-linearly warping one signal by stretching and shrinking it. Unfortunately, searching through all “warps” of a signal to find the best corresponding regions is computationally expensive. The FastDTW algorithm improves performance, but sacrifices accuracy by only considering small signal warps. 

My goal is to improve the speed of DTW while maintaining high accuracy. My key insight is that in any particular application domain, signals exhibit specific types of variation. For example, the accelerometer signal measured for two different people would differ based on their stride length and weight. My system, called Machine Learning DTW (MLDTW), uses machine learning to learn the types of warps that are common in a particular domain. It then uses the learned model to improve DTW performance by limiting the search of potential warps appropriately. My results show that compared to FastDTW, MLDTW is at least as fast and reduces errors by 60\% on average across four different data sets. 
These improvements will significantly impact a wide variety of applications (e.g. health monitoring) and enable more scalable processing of multivariate, higher frequency, and longer signal recordings.

\end{abstract}

%

%
\maketitle

\section{Introduction}
\label{sec:intro}

\begin{figure}[t]
    \begin{subfigure}{0.47\linewidth}
      \centering
      \includegraphics[width=\textwidth]{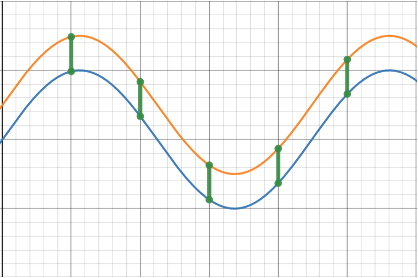} 
      \caption{Comparing Speaker 1 saying "OK, Google" against a recording}
      \label{fig:hello1}
    \end{subfigure}
    \hfill
    \begin{subfigure}{0.47\linewidth}
        \centering
        \includegraphics[width=\textwidth]{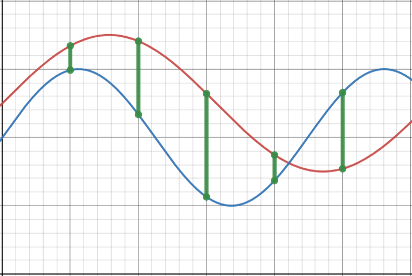} \\  
        \caption{Comparing Speaker 2 saying "OK, Google" against a recording}
        \label{fig:hello2}
    \end{subfigure}
  \caption{A sample comparison of two time series}
  \label{fig:hello}
\end{figure}

Computers collect vast amounts of information about the real-world in the form of signals. While audio might be the most obvious such signal, computers also collect signal information in the form of accelerometer readings, video, and other sensor data. A critical step in making use of these signals in applications is the ability to compare or classify signals. For example, consider a situation where you would want to do speech recognition. A device may want to determine how similar the current audio signal is to a pre-recorded sample of the phrase “OK, Google”. A simple na\"ive approach would to measure the difference between the current audio and the recording at multiple points along the signal. This difference, called the {\em Euclidean distance}, is shown in Figure~\ref{fig:hello} as green bars. 

This approach works well when the speaker's voice matches the recording closely and the sum of the resulting distances are small, identifying a match (Figure~\ref{fig:hello1}). However, speakers typically talk with different speed, volume, and tone. As a result, a second speaker saying "OK, Google" more slowly may produce a signal with a similar shape but stretched out (Figure~\ref{fig:hello2}). The end result is that measured distances would be much larger; therefore, it would fail to identify a match. The key challenge is to create a comparison mechanism that identifies signals with similar {\em shape} and ignores the variations that are common to signals in the real world.  
The ability to compare signals accurately is critical to enabling a wide range of applications including speech, gesture, handwriting, and activity recognition.

\begin{figure}[t]
  \centering
      \includegraphics[width=.75\linewidth]{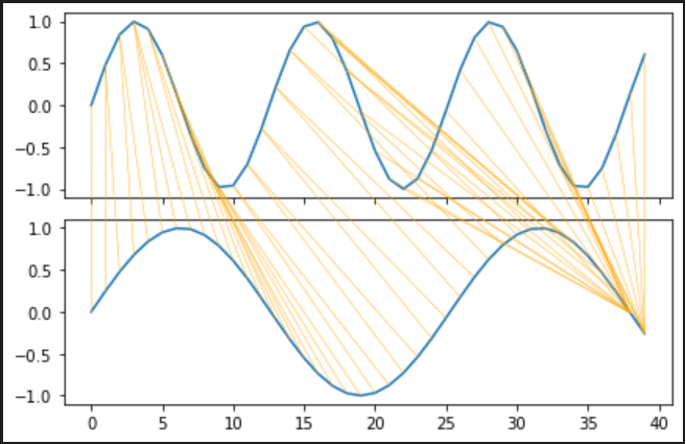} \\  
  \caption{Dynamic Time Warping between two curves.}
  \label{fig:dtw}
\end{figure}

A popular approach to quantifying the similarity between signals is to use an algorithm called {\em Dynamic Time Warping} or DTW~\cite{Sakoe-Chiba, DTWSpeech, DTWOverview}. DTW accounts for the distortions between signals by stretching and compressing one of the signals to minimize the Euclidean distance measurement. In essence, DTW finds the optimal alignment between a pair of signals that minimizes the difference between the signals. Figure~\ref{fig:dtw} shows an example of the result of the DTW alignment process. Each point on the lower signal is matched with a corresponding point on the upper graph. The mapping of points shows that the lower signal should be compressed significantly to match up with the first one and one-half cycles of the upper signal. I describe the operation of the DTW algorithm in greater detail in Section~\ref{sec:background}.

While DTW is effective at quantifying the similarity between two signals, it suffers from high computational complexity. The algorithm must search through all possible warpings (i.e., stretches and compressions) of a signal in order to find the optimal alignment. This search algorithm runs in $\mathcal{O}(n^2)$ where $n$ is the length of a signal. As a result, doubling the size of the compared signals causes the computation time to quadruple. This makes it impractical to compare signals from sensors that produce data rapidly (e.g. high-quality audio) or signals with a long duration signals (e.g., a long phrase). The most common mechanism to improve performance is to limit the possible warpings to small stretches and compressions. This approach, commonly called FastDTW~\cite{FastDTW}, does improve performance, but often at the expense of accuracy since it is not able to identify the best possible match. The goal of this paper is to develop a design that is able to improve performance over DTW while maintaining high-accuracy.

The key insight that motivates my design is that there are a common set of warps for any practical application. For example, for detection of a person walking, common factors that affect the signal are stride length and walking speed. My system, called Machine Learning DTW (MLDTW), leverages this insight to predict a likely warp path based on a small set of measurements. This predicted path is used to limit the search for the optimal warp. 

My experimental results show that MLDTW provides significant performance gains over both DTW and FastDTW. On a synthetic data set created with a range of common perturbations (noise, frequency distortion, etc.), \sys's processing time is 87.4\% less than DTW and comparable to FastDTW's time. In addition, in comparison to FastDTW, MLDTW's results had 55.8\% less error. On real-world data, \sys also performs significantly better with up to 87.3\%  less error (described in further detail in Section~\ref{sec:results}).






The rest of this paper is organized as follows. Section~\ref{sec:background} provides an overview of existing dynamic time warping algorithms and applications that use dynamic time warping. In Section~\ref{sec:system}, I describe some of the insights that lead to the design of \sys and provide an overview of how \sys operates. I then describe the design of \sys in Section~\ref{sec:software} and its implementation in Section~\ref{sec:implementation}. Section~\ref{sec:results} provides detailed results of an experimental evaluation of \sys on both synthetic and real-world data. Discussion and Conclusions are presented in Sections~\ref{sec:discussion} and~\ref{sec:conclusion}. Code associated with the implementation is available in Appendix~\ref{sec:code}

\section{Background and Related Work}
\label{sec:background}

In this section, I describe the DTW and FastDTW algorithms in detail. I also describe some common applications of DTW and why the algorithm is used. 

\subsection{DTW}
As mentioned earlier, dynamic time warping (DTW)~\cite{Sakoe-Chiba, DTWSpeech, DTWOverview} is an algorithm that is used to measure the similarity between time series data. This is achieved by finding the optimal alignment between the sequences which minimizes the Euclidean distance. It does this by finding corresponding points between the two time signals without reordering the points on the graph in any way.  

Consider two graphs of signals $A$ and $B$ with $i$ and $j$ points respectively. I refer to these points as $a_{(n_1)} \in A$ with $(1 \leq n_1 \leq i)$ and $b_{(n_2)} \in B$ with $(1 \leq n_2 \leq j)$.  In a valid alignment between signals, every point on $A$ must have at least one corresponding point on $B$ and vice versa. I refer to this property as {\em Matching}. Since every point has at least one match, there are $k \geq \max(i, j)$ such pairs. I refer to these pairs as $\{p_1,...,p_k\}$, where $p_m$, designated by $(x_m,y_m)$, implies a match between $a_{(x_m)}$ and $b_{(y_m)}$ with the constraints $(1 \leq x_m \leq i)$ and $(1 \leq y_m \leq j)$.



Each $p_m$ corresponds to one of the lines connecting the signals in Figure~\ref{fig:dtw}. The lines in Figure~\ref{fig:dtw} cannot cross each other to ensure that the warping of the signal consists only of stretching and compression without any reordering. This implies that for any two pairs $p_n$ and $p_m$, if $m > n$ then we must have $x_m \geq x_n$, $y_m \geq y_n$, and $p_n \neq p_m$ or the two corresponding lines would cross or be identical. I refer to this property as {\em Monotonicity}~\cite{DTWOverview}. 

By combining the Matching and Monotonicity properties, It can be shown that $p_1 = (1,1)$ and $p_k = (i, j)$ using proof-by-contradiction. Consider $p_1 \neq (1,1)$. This implies that $p_1 = (x_1, y_1)$ where at least one of $x_1 > 1$ or $y_1 > 1$ is true. Let us assume $x_1 > 1$ with no loss of generality. The Monotonicity property implies that $p_s$ for all $s > 1$ have $x_s > 1$. Therefore, there is no $p_s$ with $x_s = 1$ and point $a_1$ is left unmatched. This contradicts the Matching property. Since $p_1 \neq (1,1)$ is not possible in a valid alignment, we can conclude that $p_1 = (1,1)$. A similar approach (omitted for space) can be used to prove that $p_k = (i, j)$. I refer to this property as the {\em Boundary Conditions} for DTW. 

The Matching and Monotonicity properties can also be combined to constrain the relationship between $p_n$ and $p_{n+1}$.
Specifically, if $p_n = (x_n,y_n)$, then $(x_{n+1}, y_{n+1})$ for $p_{n+1}$ must be equal to one of $(x_n, y_n+1)$, $(x_n+1, y_n)$, or $(x_n+1, y_n+1)$. I use a proof-by-contradiction approach to show this property holds. Imagine that $p_{n+1}$ is not one of the above three values; the Monotonicity property will require that at least one of $x_{n+1} \geq x_n + 2$ or $y_{n+1} \geq y_n + 2$ is true. I assume $x_{n+1} \geq x_n + 2$ with no loss of generality. However, this implies that there is no $p_s$ with $x_s = x_n+1$ and no possible match for $a_{(x_n+1)}$, which contradicts the Matching property. This proves that $p_{n+1}$ must be one of $(x_n, y_n+1)$, $(x_n+1, y_n)$, or $(x_n+1, y_n+1)$. I refer to this as the {\em Step} property.


    







I can use the above properties to compute the DTW matching between two graphs $A$ and $B$, assuming I already know the optimal matching for subgraphs of $A$ and $B$. I use the expression $Dist(s, t)$ to refer to the Euclidean distance between points $a_s$ of signal $A$ and $b_t$ of signal $B$. I use the expression $D(x,y)$ to represent the total Euclidean distance or cost associated with the optimal matching between the first $x$ points of signal $A$ and the first $y$ points of signal $B$. Recall that signal $A$ and $B$ have $i$ and $j$ points respectively and that point $i$ must correspond to point $j$ because of the Boundary Conditions property. This indicates that $p_k = (i, j)$ is part of the optimal DTW matching, meaning that $Dist(i, j)$ is part of the total resulting optimal DTW distance. The pair $(i, j)$ is effectively the right-most line in a matching such as the one shown in Figure~\ref{fig:dtw}. The Step property implies that the next line (i.e., $p_{k-1}$) must be $(i-1, j), (i, j-1)$ or $(i-1, j-1)$. It is possible to compute the resulting total distance associated with each of the three choices. Once computed, DTW can simply use the minimum resulting total cost. Using this observation, I can express the total DTW distance as:

\begin{equation}
\begin{aligned}
    D(i, j) = {} & Dist(i, j)  {~} + \\
    & min(D(i-1, j), D(i, j-1), D(i-1, j-1))
\end{aligned}
\label{eq:dtw}
\end{equation}

The above recurrence relationship can be used to implement a recursive solution to computing the DTW matching. Recall that the Boundary Conditions property also required $p_1 = (1,1)$. As a result, $D(1,1) = Dist(1,1)$. This can be used to terminate the recursion and enable the computation of $D(x,y)$ for any $x$ and $y$. Once the minimum distance is computed, I can use the computation to determine the matching between points (i.e., the warp path). Each $Dist(x,y)$ used in computing $D(i,j)$ (i.e., the values chosen by the min() in Equation~\ref{eq:dtw}) is associated with a single pair of corresponding points in the warp path. I refer to this step as the {\em backtracking} phase of DTW.

Note that computing $D(x,y)$ requires computing the values of all the $D(x,y)$ values along the warp path. However, since it is impossible to know the warp path in advance, the DTW algorithm must compute $D(a,b)$ for all $a<x$ and all $b<y$. A na\"ive approach to using the recurrence would result in a computation that takes $\mathcal{O}(3^n)$ time, assuming $x=y=n$. This is impractical for anything other than very small $n$. However, a dynamic programming approach could optimize the recursion by caching previous computations of $D()$. In essence, each $D(a,b)$ would be computed exactly once with a relatively simple computation. Since, $a$ can take on values $0$ to $x$ and $b$ can take on values $0$ to $y$, there are $x \times y$ computed values. This results in a computational complexity of $\mathcal{O}(n^2)$ assuming that $x$ and $y$ are both equal to $n$.

\begin{figure}[t]
  \centering
      \includegraphics[width=.75\linewidth]{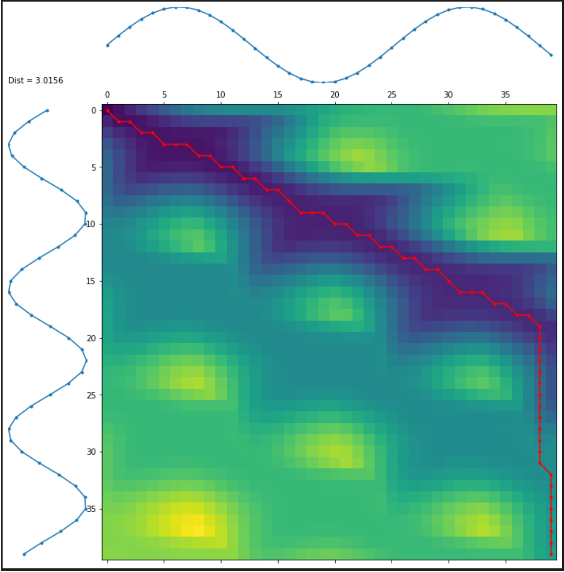} \\  
  \caption{Sample cost matrix for two sine waves. The red line represents the warp path. Dark colors represent small values for $D(i,j)$ and light colors represent large values.}
  \label{fig:dtwmatrix}
\end{figure}

Figure~\ref{fig:dtwmatrix} illustrates the results of this computation for two sine waves. Each square in the matrix, called a {\em Cost or Distance Matrix}, represents a single $D(a,b)$ value. The bottom right entry represents the total distance for matching the two signals. The red line represents the warp path and is computed by backtracking from the bottom right entry and following the matrix entries that it depends on. 

A major drawback with using DTW is that it is slow. In the next section, I describe a common optimization to improve performance, FastDTW~\cite{FastDTW}. I use the name {\em FullDTW} to refer to the baseline DTW algorithm described above and to differentiate it from the DTW variants that I describe in the rest of the paper.

\subsection{FastDTW}
\label{sec:fastdtw}

FullDTW's slow speed is a result of having to compute $D(x,y)$ for all possible values of $x$ and $y$ for a pair of signals. In reality, the FullDTW computation only relies on the few $D(x,y)$ along the true warp path. FastDTW~\cite{FastDTW} is built on the observation that computing $D(x,y)$ for all possible values of $x$ and $y$ may not be necessary. Imagine computing $D(x,y)$ for a small subset of potential $(x,y)$ values and just assigning $\infty$ to $D(x,y)$ for all $(x,y)$ that were not chosen. If the backtracking algorithm was used to compute a warp path, the resulting path would be constrained to only the chosen $(x,y)$ values.

The FastDTW algorithm adopts this approach, choosing a simple geometrically defined subset of values such as those near the diagonal, also called the Sakoe-Chiba Band~\cite{Sakoe-Chiba}, of the cost matrix. Note that other statically defined shapes, such as Itakura Parallelogram~\cite{Itakura}, have also been used for FastDTW. These shapes are defined by a fixed constant. As a result, the number of distance entries that must be computed increases linearly with the number of points in the signal. This makes FastDTW run in time $\mathcal{O}(n)$, which results in much better scalability than FullDTW.

Note that, in practice, using a dynamic programming approach, it is not possible to precisely compute $D(x,y)$ for a subset of $(x,y)$. Recall that $D(x,y)$ any particular $(x,y)$ depends on $D(x', y')$ for all $x' < x$ and $y' < y$. However, since not all values are computed, the computation of $D(x,y)$ can only rely on the values that are computed. As a result, the FastDTW approach suffers from inaccuracies from two sources. First, its warp path is limited to the range of computed distance values, and second, the distance values, themselves, may be inaccurate since they are also limited to depend only on the computed subset of distances. 

\begin{figure}[t]
    \begin{subfigure}{0.47\linewidth}
      \centering
      \includegraphics[width=\textwidth]{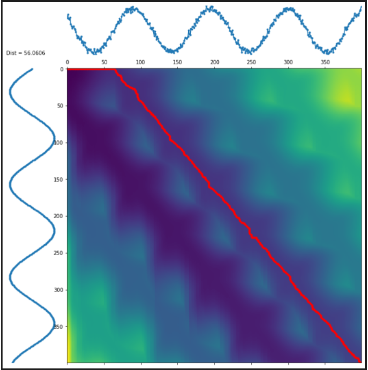} 
      \caption{DTW: Highly Accurate, but Slow}
      \label{fig:DTWFullcompare}
    \end{subfigure}
    \hfill
    \begin{subfigure}{0.47\linewidth}
        \centering
        \includegraphics[width=\textwidth]{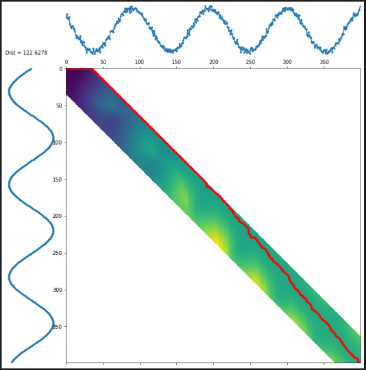} \\  
        \caption{FastDTW: Fast, but Inaccurate}
        \label{fig:DTWFastCompare}
    \end{subfigure}
  \caption{Comparison between FullDTW and FastDTW on two sine waves. FastDTW(b) fails to find the optimal path due to search constraints in the top left portion of the graph.}
  \label{fig:compare}
\end{figure}

Figure~\ref{fig:compare} illustrates the difference between FullDTW and FastDTW. As before, each colored point in the figure represents a single $D(x,y)$ value (lighter color indicates higher distance and vice versa) and the red line represents the warp path. In the case of FullDTW (Figure~\ref{fig:DTWFullcompare}), every single possible $D(x,y)$ is computed and the red line is the optimal possible warp path. In the case of FastDTW (Figure~\ref{fig:DTWFastCompare}), $D(x,y)$ is not computed for many points, represented by the white region in the graph. In this particular configuration, FastDTW only computes distances close to the diagonal. The resulting warp path is constrained to the colored region, although the optimal path, shown in Figure~\ref{fig:DTWFullcompare}, would clearly be in the white portion of the graph. Also, note that the coloration of graph differs between FullDTW and FastDTW.

Constraining the search space is not the only way to improve the performance of FullDTW. Two other common techniques are data abstraction, and indexing~\cite{FastDTW}. Data abstraction involves running DTW on a reduced representation of the data, essentially decreasing the resolution of the cost matrix. Indexing prunes the number of times DTW needs to run in certain tasks by using lower-bounding functions. This includes methods such as clustering sets of times series in order to predict the likely distance for a time series based on the distance computed for a similar time series. In this paper, I focus on search space constraints since data abstraction and indexing can be used in conjunction with my method to provide additional performance gains.

\subsection{Example Applications}

DTW is used in a variety of applications that require the ability to compare two signals. Here, signals simply refer to readings or observations that have some type of ordering, usually due to time. For example, it can be used in healthcare monitoring for early diagnosis of cardiac arrhythmia ~\cite{DTWMedicine} and in home applications such as home energy consumption monitoring~\cite{DTWEnergy}. To provide a concrete example of how DTW is used in practice, I describe its use in handwriting recognition~\cite{DTWHandwriting} below. 

Consider a modern touch screen or tablet. As a user moves the pen along the surface, it produces a sequence of $(x,y)$ coordinates of the pen over time. If an application wants to identify a drawn symbol as a letter or gesture, it needs to compare the sequence of coordinates with recorded examples of letters and gestures. Unlike the earlier examples used, such as audio recognition, each point in this signal has two values: $x$ and $y$. However, a DTW algorithm comparing points on the signals still relies on a distance measurement between points. In the case of audio signals, the distance is simply the difference in the magnitude of the signals. However, for handwriting, the Euclidean distance between the measured signal and reference coordinates needs to be computed, i.e., $\sqrt{(x_{signal} - x_{reference})^2 + (y_{signal} - y_{reference})^2}$

\begin{figure}[t]
    \begin{subfigure}{0.47\linewidth}
      \centering
      \includegraphics[width=\textwidth]{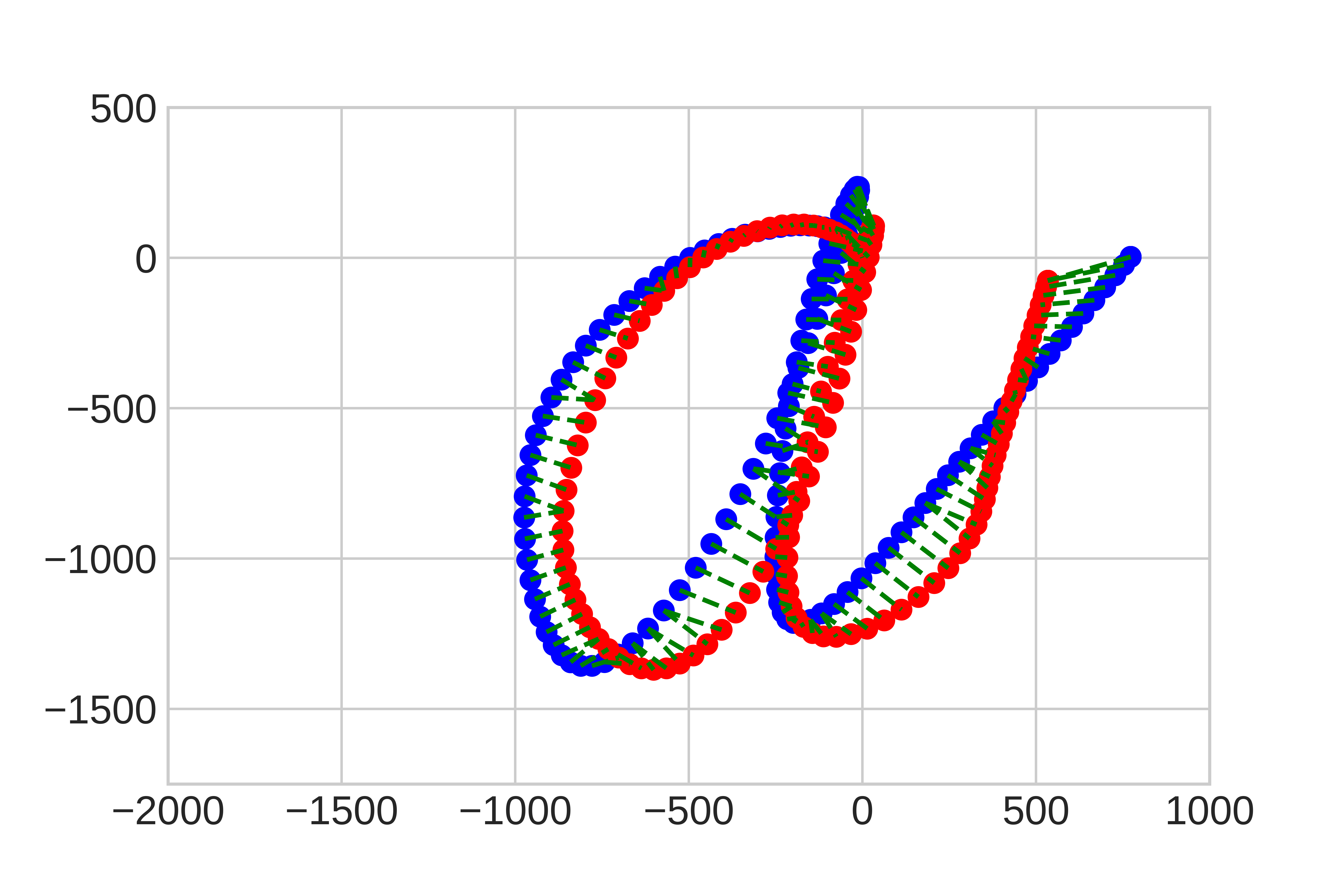} 
      \caption{DTW for letter a - distance 23095}
      \label{fig:HWaa}
    \end{subfigure}
    \hfill
    \begin{subfigure}{0.47\linewidth}
        \centering
        \includegraphics[width=\textwidth]{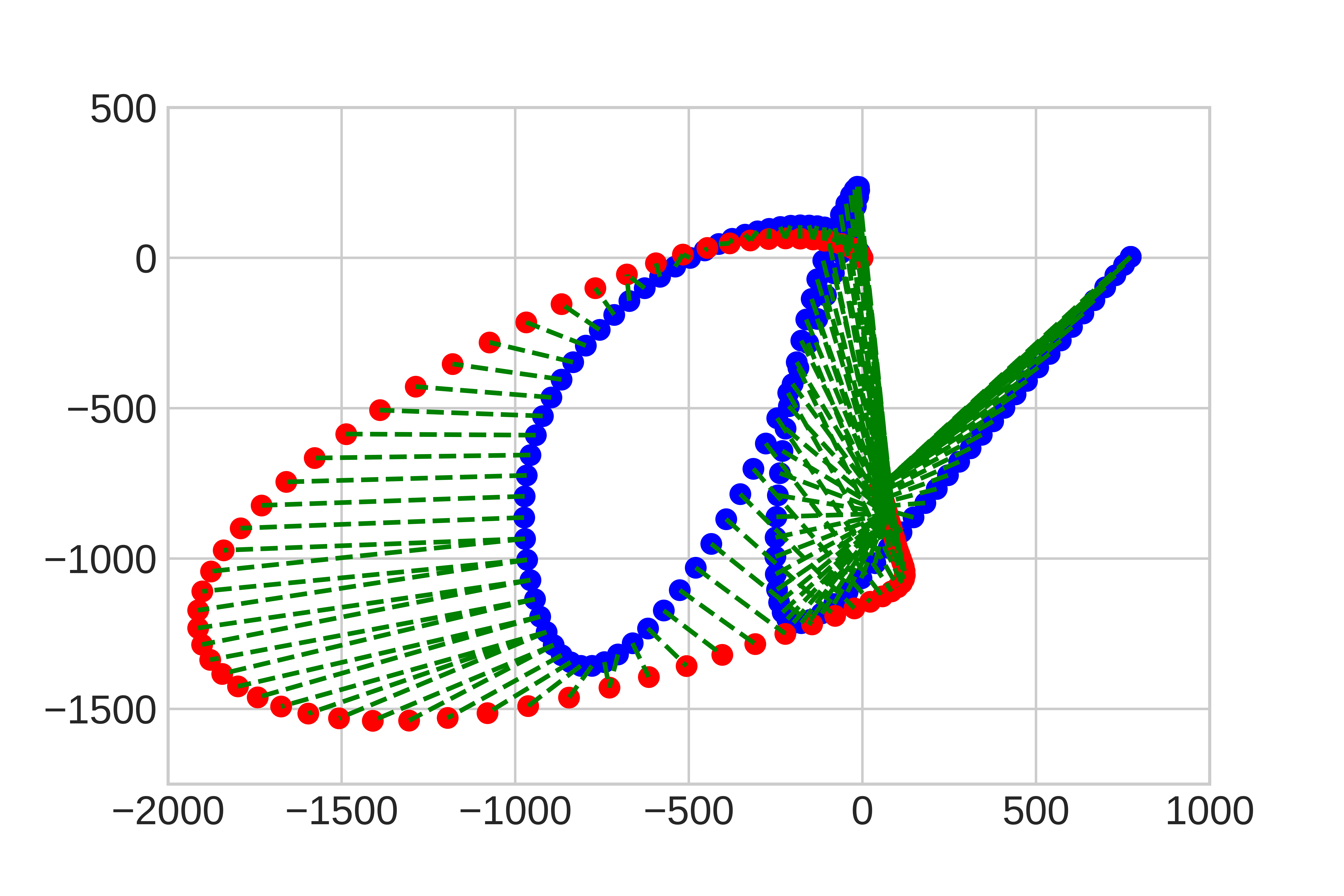} \\  
        \caption{DTW for letter a to c - distance 110268}
        \label{fig:HWac}
    \end{subfigure}
  \caption{DTW can be used to compare letters written by different writers to track forgery. The DTW distances shows that the letter drawn in blue is far more similar to pre-recorded "a" than a "c" (shown in red). The warp path is shown in green. }
  \label{fig:HW}
\end{figure}

Figure~\ref{fig:HW} illustrates the result of using DTW to recognize letters. The blue dots represent the measured signal of the letter drawn by the user. The red dots in Figure~\ref{fig:HWaa} and Figure~\ref{fig:HWac} represent the reference measurements for the letters a and c, respectively. The green lines represent the best warp path computed by DTW. The DTW distance for Figure~\ref{fig:HWaa} is 23,095 while the DTW distance for Figure~\ref{fig:HWac} is 110,268, showing that the drawn symbol is a much closer match to the reference letter a than to c. In addition to identifying letters, this approach could also be used to detect fraud by comparing the captured handwriting with an authorized user's pre-recorded handwriting or signature. I have used the DTW designs described in this paper to implement a signature recognition application, which is described in greater detail in Section~\ref{sec:discussion}.

\section{Insights and Challenges}
\label{sec:system}

As mentioned in Section~\ref{sec:fastdtw}, the source of FastDTW error is that the subset of distances  it chooses to compute may not include the optimal warp path. One factor that contributes to this issue is that FastDTW uses a fixed subset of potential warps that it considers across all instances. My first insight is that there should be certain types of warps common to a specific application domain. For example, consider the application of recognizing activities based on accelerometer readings from a smart wristband, watch, or phone. This application works by having signature accelerometer readings for walking, running, bicycling and other common activities. However, there are common factors that affect these signatures such as height, stride length and walking speed. In addition, it should be possible to identify the impact of these factors on warp paths by analyzing large data sets of signals recorded from an application. For example, analyzing a handwriting data set would help identify the range of different ways that people draw the letter "a" (see Figure~\ref{fig:HWaa}). 

My second insight is that it should be possible to identify the factors affecting a particular recording from a small sample of the overall signal. In the case of an application such as activity recognition, it should be possible to estimate the height or speed of a person from a subset of the overall signal. Even if you cannot estimate factors such as height or speed directly, it should be possible to estimate the impact that these factors are having on the optimal warp path. 

\begin{figure}[t]
  \centering
      \includegraphics[width=.75\linewidth]{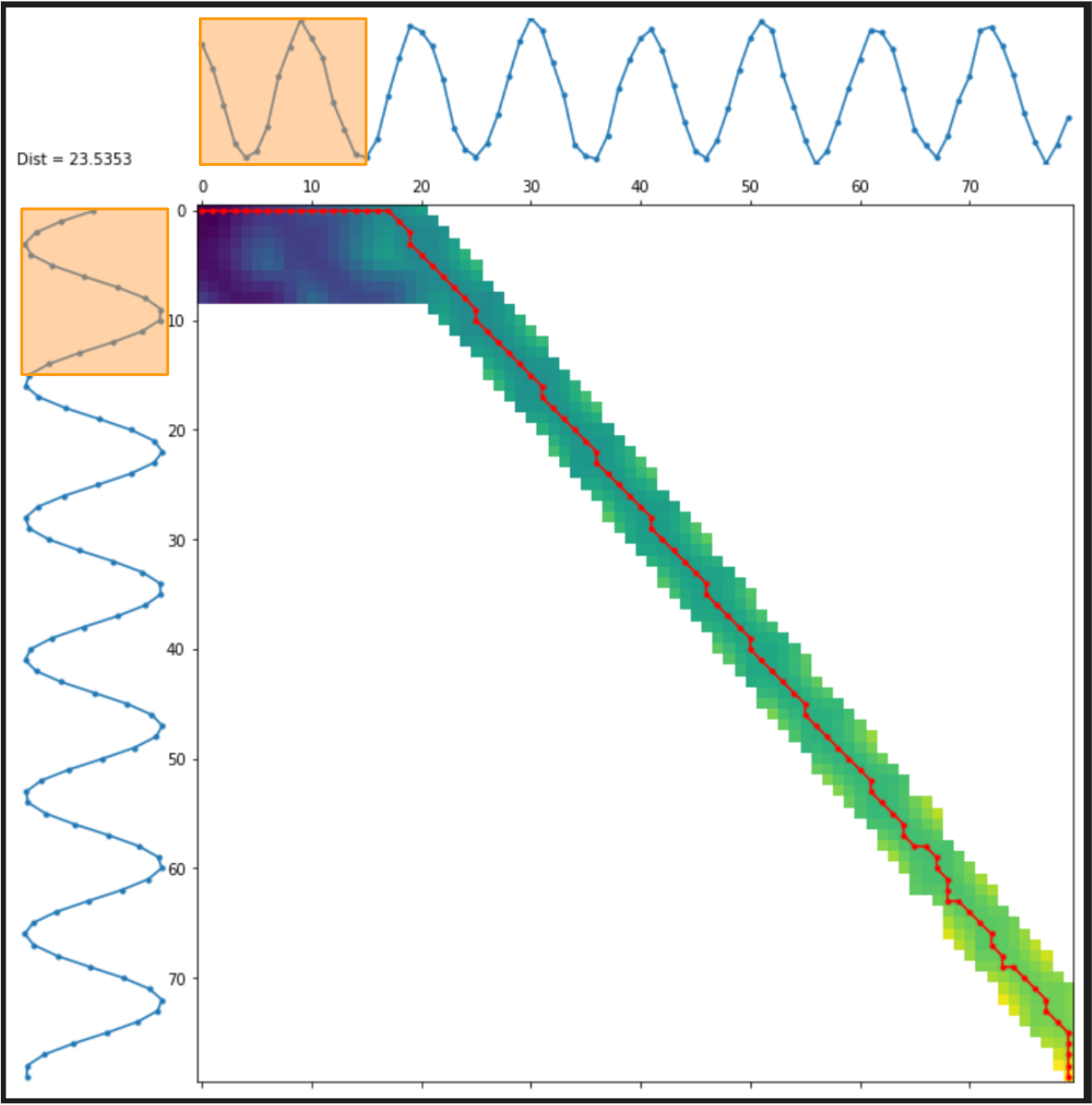} \\  
  \caption{MLDTW uses a small subset of the signal values (highlighted in orange) to predict the best part of the cost matrix to search for the optimal warp path.}
  \label{fig:mldtw}
\end{figure}

These two insights led me to the design of Machine Learning DTW (MLDTW). The basic concept behind this design is illustrated in Figure~\ref{fig:mldtw}. \sys begins by taking a small subset of each time series -- highlighted by orange boxes in the figure. The algorithm uses the subset of values provided to predict a potential warp path. A region around this predicted warp path is used to account for some uncertainty and then distance values are calculated only within this region. After this point, the normal DTW approach to compute the warp path using distance measurements is employed. If the warp region is accurately estimated, as shown in the figure, the resulting warp path will be close to optimal. Note that FastDTW would have picked the direct diagonal of the square matrix shown. This would have been far from the optimal warp path, resulting in significant error. 

The key step in this ideal design is generating the warp path prediction. As its name suggests, MLDTW uses machine learning models for this purpose. In the next section, I describe in detail how these models are trained and used. 

\section{\sys Design}
\label{sec:software}

In this section, I provide an overview of my MLDTW design. 

\subsection{Overview}

MLDTW goes through the following steps to make the necessary predictions. First, it collects the first $n$ values of each time series. These are the {\em features}, or what will be used to determine the likely location of the ideal warp path. The prediction of the warp path is produced in the form of 5 equally spaced waypoints. This approach is illustrated in Figure~\ref{fig:mltrain}. The predicted warp path is created by linking the waypoints into a piece-wise linear shape. The use of waypoints reduces the level of detail of the prediction since it does not make a prediction for every point along the warp path. However, waypoints provide a practical balance between precision of the prediction and complexity of the machine learning model needed. 

One additional step is needed to convert the predicted warp path into a search space for the MLDTW algorithm. A simple approach would be to have a fixed width search space around the predicted path, similar to the Sakoe-Chiba Band commonly used in FastDTW implementations. However, I can allocate the search space more accurately based on my prediction mechanism. The machine learning model also provides a confidence value associated with each of the waypoints. Since this indicates how likely a particular point in the path is on the true optimal warp path, I can use high confidence to shrink the search space and low confidence to expand the search space. Based on this observation, the width of the search area dynamically adjusts based on the confidence levels of each waypoint prediction, following an inverse relationship.

\subsection{Machine Learning Architecture}

\sys uses a neural network classifier to predict the waypoints. I tested several different types of machine learning models before deciding to use a neural network including a random forest model and a support vector machine. Neural networks were superior for multiple reasons. First, there is no linear trend to the data, making algorithms such as support vector machines inadequate. In addition, there are a large number of features and I want to prioritize accuracy over training time.

Two different network architectures were designed. The two designs treat the situation as a classification and regression problem respectively. As both achieved similar performance, code segments and results are only provided for one of the two in the remainder of the paper. In general, the paper presents the Classification Approach unless otherwise noted. 

\subsubsection{Classification Approach}
\begin{figure}[t]
    \begin{subfigure}{0.47\linewidth}
      \centering
      \includegraphics[width=\textwidth]{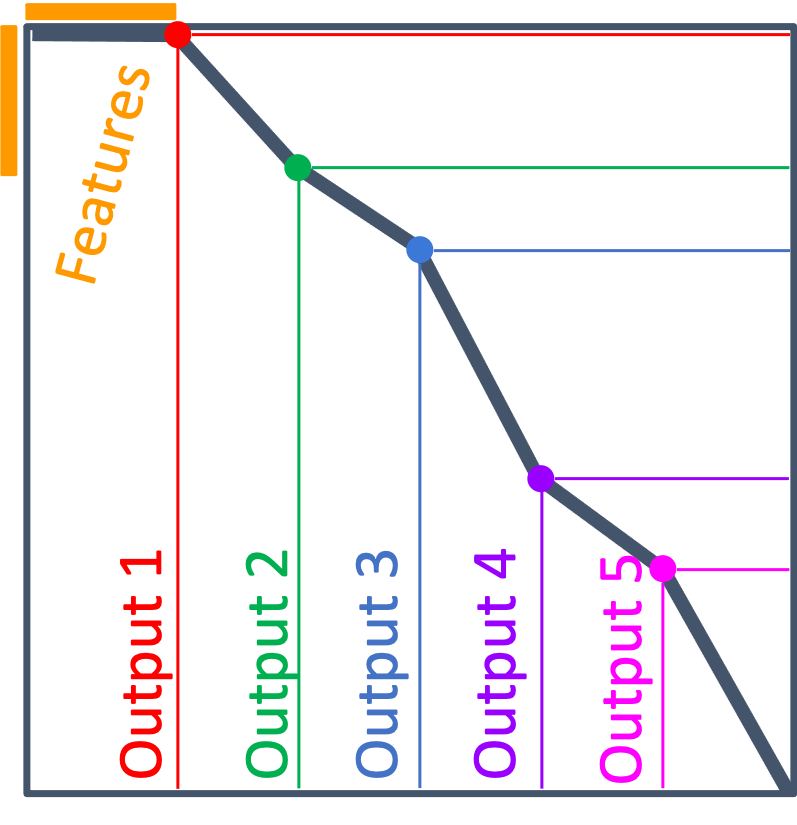} 
      \caption{MLDTW's uses the starting subset of each time series labeled as features to predict locations of waypoints.}
      \label{fig:mltrain}
    \end{subfigure}
    \hfill
    \begin{subfigure}{0.47\linewidth}
        \centering
        \includegraphics[width=\textwidth]{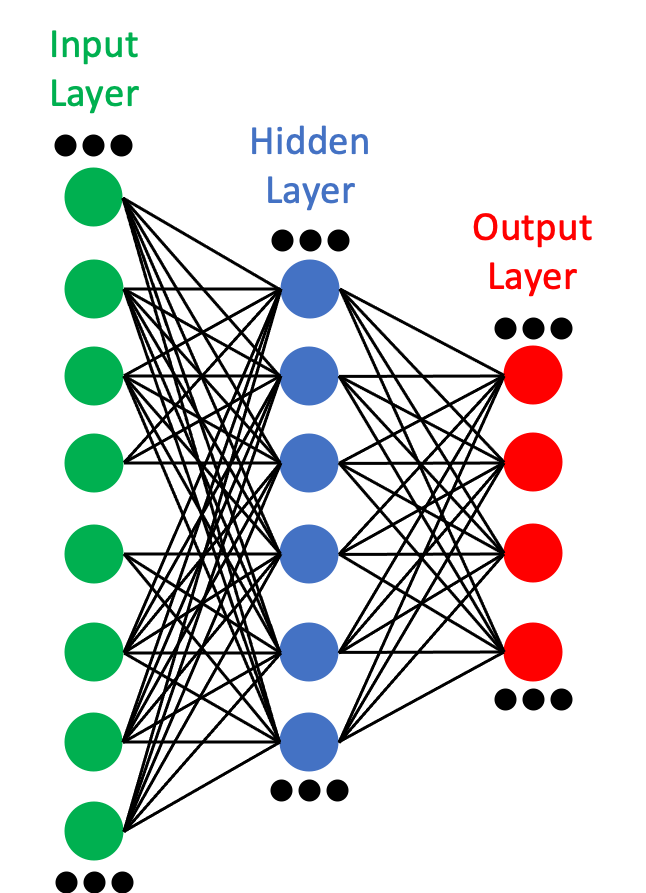} \\  
        \caption{Neural networks consist of three parts: an input layer, optional hidden layers, and an output layer, each of which each contain neurons }
        \label{fig:layers}
    \end{subfigure}
  \caption{MLDTW neural network}
  \label{fig:mlsys}
\end{figure}

In the classification design, the waypoints are the {\em labels}, or what the model is attempting to predict. \sys creates a neural network for each waypoint. Neural networks consist of three parts: an input layer, optional hidden layers, and an output layer, which each contain neurons (Figure~\ref{fig:layers}). The output layer must contain the same number of neurons as there are unique labels. Note that using separate neural networks for each waypoint prediction makes the network simpler and significantly reduces the number of output neurons.

Each layer also has an {\em activation function} for its neurons. An activation function decides whether a neuron should be activated on not based on its input. The two that I use are Rectified Linear Units or ReLU~\cite{relu} and Softmax~\cite{generalML}. ReLU is based on a linear model and is the most common activation function. It is defined by the following equation:
\begin{equation} \label{eq:2}
{\displaystyle \phi (\mathbf {v} )=\max(0,a+\mathbf {v} '\mathbf {b} )},
\end{equation}
Softmax is based on an exponential model and is often used in the output layer to normalize the output to a probability distribution. This is used to produce confidence scores for labels.
It is defined by the following equation:
\begin{equation} \label{eq:3}
\sigma(\vec{z})_{i}=\frac{e^{z_{i}}}{\sum_{j=1}^{K} e^{z_{j}}}
\end{equation}

\subsubsection{Regression Approach}
In the regression design, the waypoint locations are a continuous numeric output and all locations are predicted by a single model. It uses the ReLU activation function for all layers. In order to produce a confidence score, an ensemble of several models is created. Each model produces its predictions for the waypoint locations, creating distributions. From these distributions, 95\% confidence intervals can be formed for each location. This is possible due to the stochastic behavior of neural networks, such as the randomized weights that are initialized during training. This results in models trained on the same data set having slightly different outputs~\cite{mlRegression}.

\section{\sys Implementation}
\label{sec:implementation}

In this section, I describe how \sys implements the design described in Section~\ref{sec:software}. A more complete code listing associated with the different components of \sys is provided in Appendix~\ref{sec:code}.

In order to train the machine learning portion of MLDTW (Section~\ref{sec:training}), I need a labeled ground-truth data set. This requires two parts, a collection of signal measurements for a particular application and an implementation of FullDTW to provide labels for the true optimal warp path (Section~\ref{sec:dataset}). 

MLDTW must then use the trained machine learning model as part of its search for a minimum cost warp path. I describe how the implementation incorporates the machine learning model in Section~\ref{sec:usingPredictions}. 

\subsection{Data Sets}
\label{sec:dataset}

I apply \sys on 4 data sets from 3 different application domains in this study. These data sets are described below. I also describe the process of generating the features and labels that are needed for the training on these data sets.

\subsubsection{Synthetic Data}

To illustrate the capabilities of MLDTW on a general data set that features some of the common issues in signal processing, I created a synthetic data set. DTW is designed to focus on the shape of the curve and minimize the impact of noise, magnitude and speed/frequency of the signal. To see how well DTW variants handle noise, temporal offsets, and frequency variation, my synthetic data set, which I call \texttt{SYNTH}, explicitly includes these factors. The data set consists of 10,000 sine waves of different frequency with random noise equal to 7.5\% of the signal magnitude added to each signal sample. The signals were also shifted to start at an arbitrary phase. 

\subsubsection{Handwriting Recognition}

The UJI Pen Characters Data Set~\cite{UJIPenChars}, obtained from UCI Machine Learning Repository~\cite{Dua:2019} is a data set containing samples of characters drawn by 11 writers. The data set contains all the upper and lower case A-Z English alphabet plus digits. There are 1364 samples in the data set: 11 writers x 2 repetitions x (2x26 letters + 10 digits). The data consists of X and Y coordinate positions along the stroke of each character. Initial experiments were done using this data set. However, I chose to create my own character data set to provide more detailed traces. 

My data set was collected through a custom application I designed that provides a simple drawing canvas. I drew 200 repetitions of each character to test MLDTW. I call this data set \texttt{WRITING}.

\subsubsection{Activity Recognition}

The Smartphone-Based Recognition of Human Activities and Postural Transitions Data~\cite{accActivity}, obtained from UCI Machine Learning Repository set is a human activity recognition data set that provides accelerometer data from a smartphone for activities such as walking, sitting, and more. This data set is based on 30 subjects. It has 1821 time series data for x,y, and z acceleration data in meters per second squared during walking to train the 5 models needed for MLDTW. I calculate the magnitude of acceleration (i.e., $\sqrt{{a_x}^2 + {a_y}^2 + {a_z}^2}$). I call this data set \texttt{ACTIVITY}.

\begin{figure}[t]
  \centering
      \includegraphics[width=.6\linewidth]{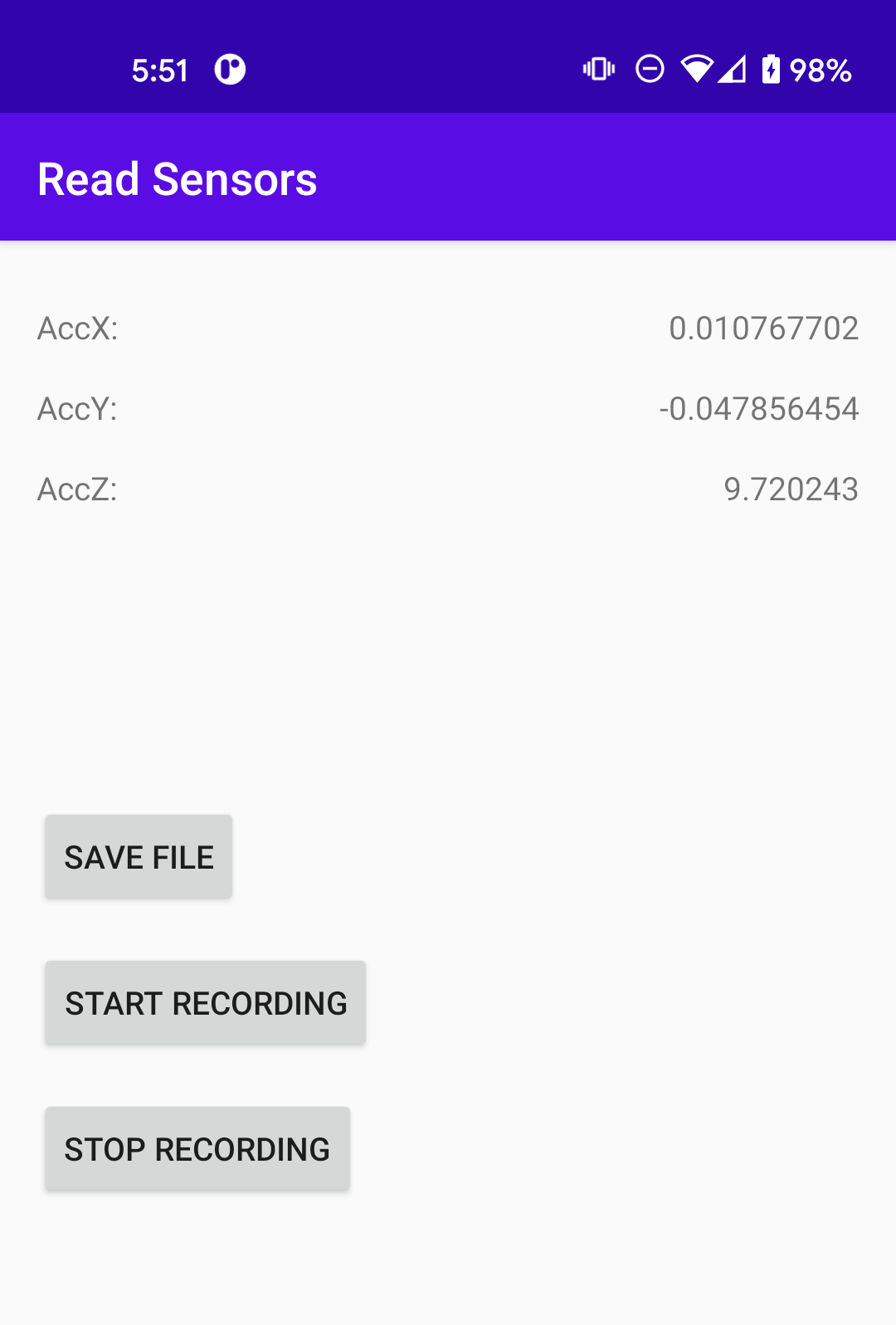} \\  
  \caption{Screenshot of Android application that collects accelerometer data for \texttt{WALKING} data set. }
  \label{fig:app}
\end{figure}

To further test the ability of MLDTW on activity recognition, I created a program to collect my own data. A screenshot of the program's interface is shown in Figure~\ref{fig:app} and code is provided in Appendix~\ref{sec:acccollectionapp}. The program collects x, y, and z acceleration readings at 200Hz. I collected measurements for two different individuals walking indoors. Again, I use the magnitude of acceleration. There are 27,659 data points total. I call this data set \texttt{WALKING}.

\subsubsection{Labeling Optimal Warp Path}
\label{sec:labels}

\begin{figure}[t]
\begin{lstlisting}[language=Python]
# create empty cost matrix
dtwMatrix = np.zeros((n+1, m+1))
for i in range(n+1):
    for j in range(m+1):
        dtwMatrix[i, j] = np.inf
dtwMatrix[0, 0] = 0
for i in range(1, n+1):
    for j in range(1, m+1):
        # calculate Euclidean distance
        cost = abs(s1[i-1] - s2[j-1])
        lastMin=np.min([
            dtwMatrix[i-1, j],
            dtwMatrix[i, j-1],
            dtwMatrix[i-1, j-1]])
        # calculate total distance for subset of time series
        dtwMatrix[i, j] = cost + lastMin
dist = dtwMatrix[len(dtwMatrix)-1][len(dtwMatrix[0])-1]
\end{lstlisting}
\caption{Code computes the full cost matrix used by DTW.}
\label{fig:dtwcostimpl}
\end{figure}

\begin{figure}[t]
\begin{lstlisting}[language=Python]
best_path = [(curRow, curCol)]
while curRow != 0 or curCol != 0:
    # Chooses the direction (diagonal, left, up) with the least distance in the cost matrix
    dirs = [paths[curRow][curCol],
        paths[curRow+1][curCol],
        paths[curRow][curCol+1]]
    if dirs.index(min(dirs)) == 2:
        curRow -= 1
    elif dirs.index(min(dirs)) == 1:
        curCol -= 1
    else:
        curRow -= 1
        curCol -= 1
    best_path.insert(0, (curRow, curCol))
\end{lstlisting}
\caption{Code finds the ideal warp path that minimizes the distance.}
\label{fig:dtwbackimpl}
\end{figure}

With each of the above data sets, I need the  a set of five waypoints along the optimal warp path and the first $n$ values of each time series as indicated in Figure~\ref{fig:mltrain}.

The Python code in Figure~\ref{fig:dtwcostimpl} creates the cost matrix. First, a matrix of infinities of the dimensions of the two time series is created. The top left value is set to 0.  Recall that the value of a matrix entry is only dependent on values above and to the left of the entry's location. This makes it possible to compute the values for each column from top to bottom, processing the columns from left to right. The two loops, for $i$ and $j$, correspond to processing the columns from left to right and a single column from top to bottom, respectively. The cost is calculated as the difference between the two points at that position. This is then added to the minimum of the cells before it, which is either to the left, up, or diagonal, to find the value at that position. Finally, the bottom right value represents the distance between the two time series.

Once I have the cost matrix computed, the code in Figure~\ref{fig:dtwbackimpl} finds the ideal warp path that minimizes the distance. It does this by using the cost matrix and backtracking. It starts at the bottom right and follows where the distance is minimized, either up, left, or diagonally towards the up and left.

With the warp path calculated, I extract the first $n$ values of each time series and five waypoints needed for training. 

\subsection{Training}
\label{sec:training}

\begin{figure}[t]
\begin{lstlisting}[language=Python]
outputNeurons0 = len(np.unique(y0))
# Employs early stopping to stop training when validation loss increases
early_stopping0 = EarlyStopping(monitor='val_loss', 
    patience=10)

# Creates model with 2 layers with ReLU and Softmax activation functions
model0 = models.Sequential()
model0.add(layers.Dense(300,  activation='relu', 
        input_shape =(X_train_ML_0.shape[1],)))
model0.add(layers.Dense(outputNeurons0,  
        activation='softmax'))
model0.compile(optimizer='adam', 
        loss='sparse_categorical_crossentropy', 
        metrics =['accuracy'])
# Trains model
history0 = model0.fit(X_train_ML_0,
        y_train_ML_0,
        epochs=200,
        validation_data =(X_test_ML_0, y_test_ML_0),
        callbacks=[early_stopping0])
\end{lstlisting}
\caption{Sample code for creating and training one of the five neural networks.}
\label{fig:training}
\end{figure}

\begin{figure}[t]
  \centering
      \includegraphics[width=.8\linewidth]{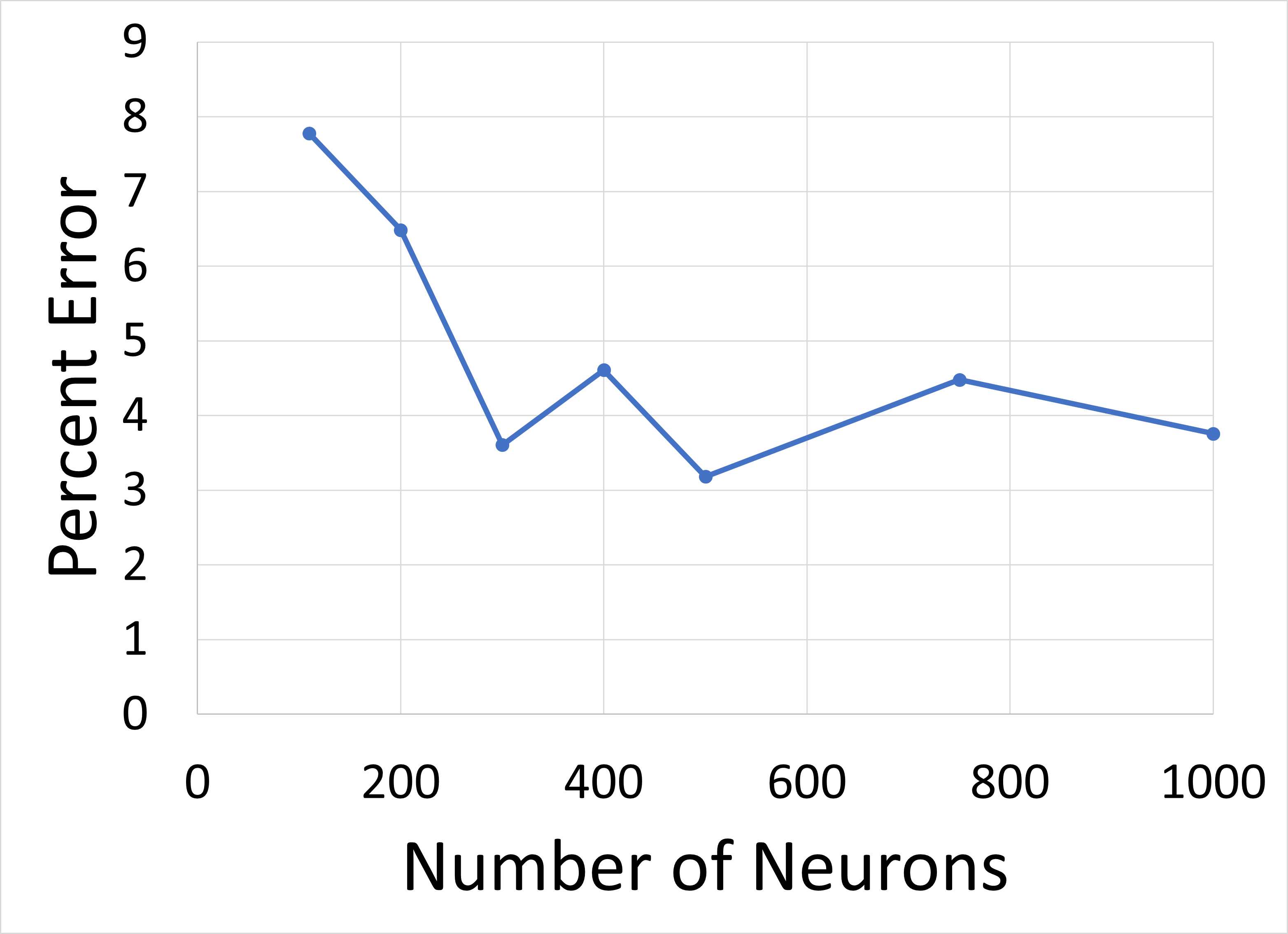} \\  
  \caption{Impact of neural network size on error.}
  \label{fig:neuronError}
\end{figure}

\begin{figure}[t]
    \centering
    \begin{subfigure}{0.8\linewidth}
        \includegraphics[width=\textwidth]{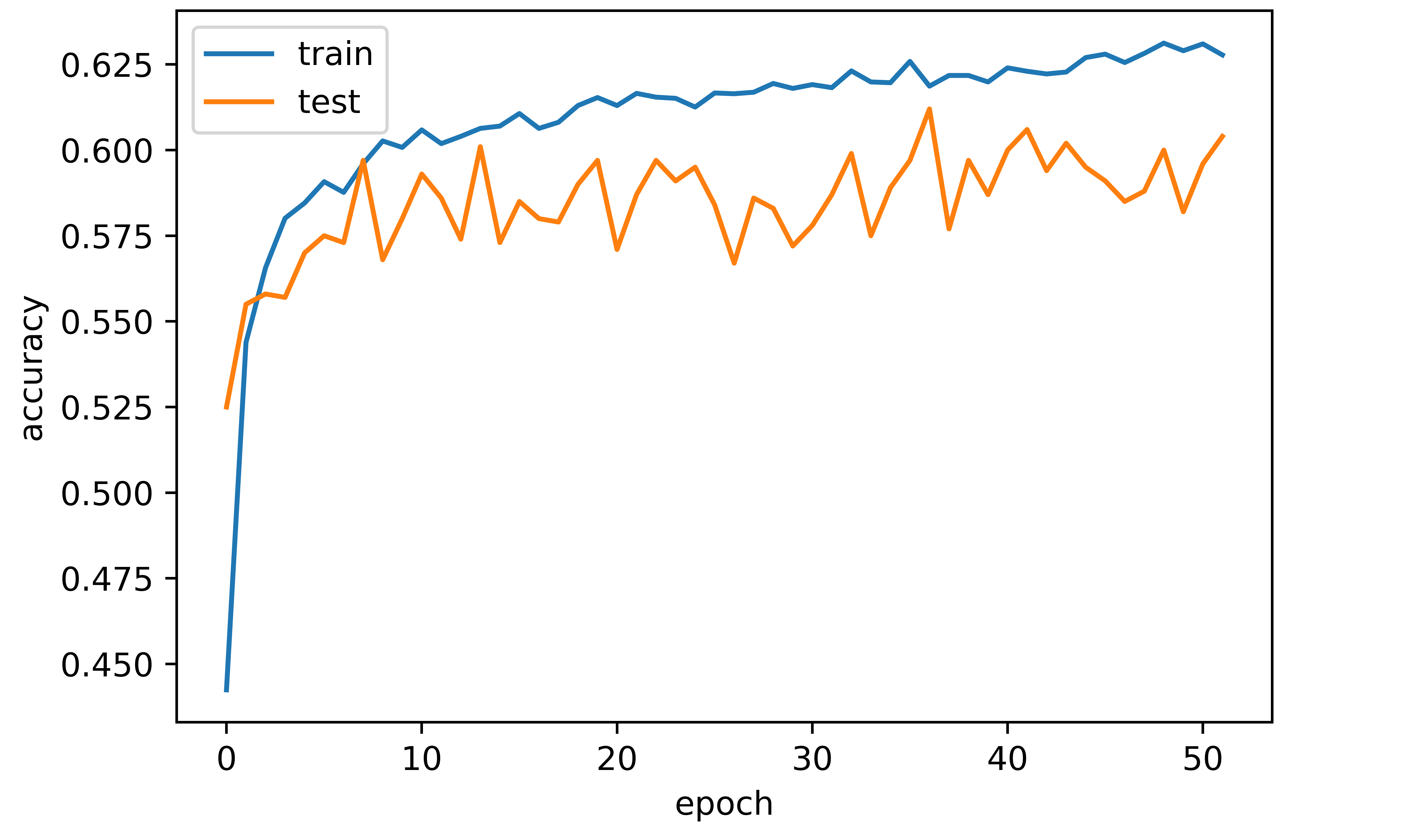} \\  
        \caption{Model Accuracy}
        \label{fig:modelAcc}
    \end{subfigure}
    \hfill
    \begin{subfigure}{0.8\linewidth}
        \includegraphics[width=\textwidth]{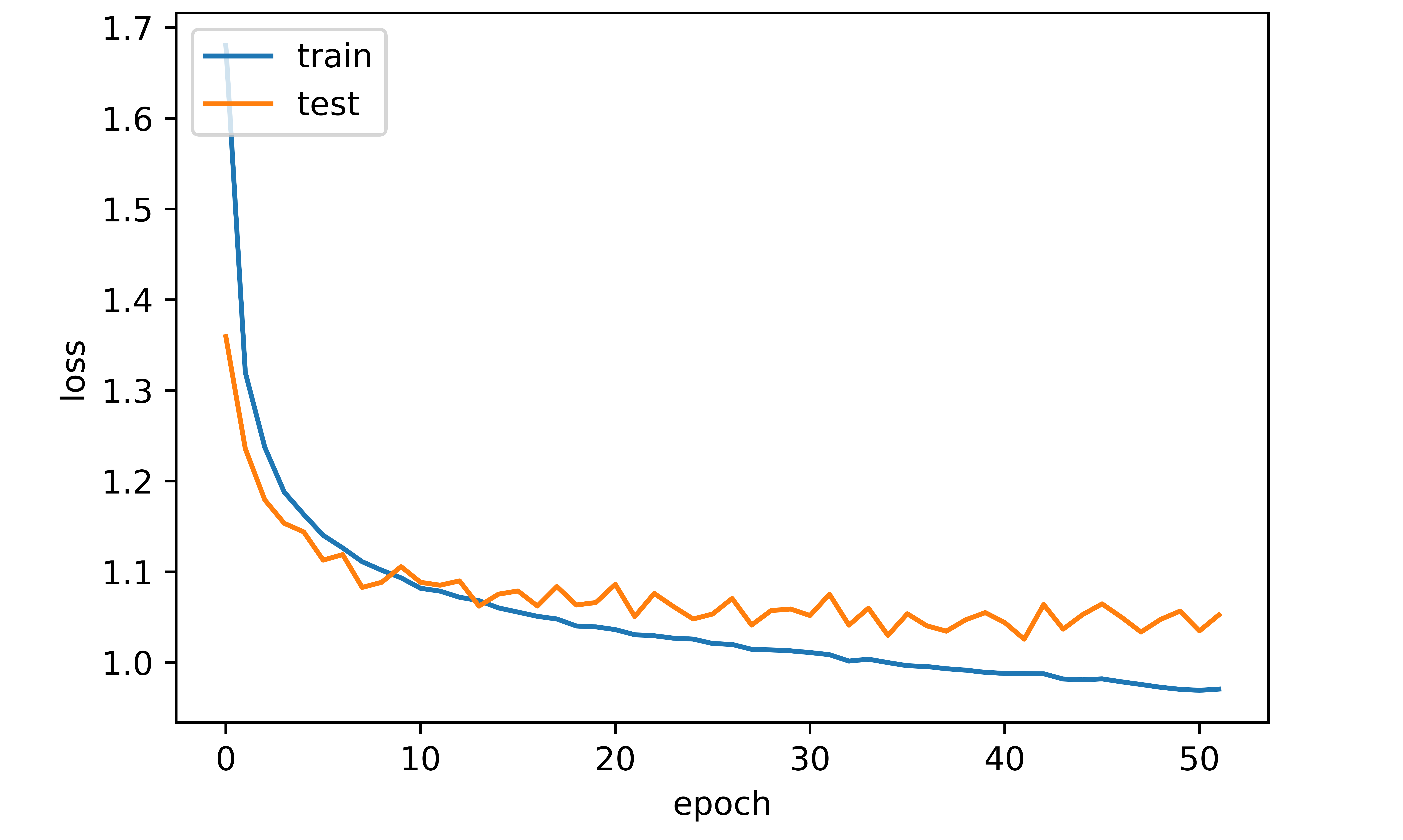} \\  
        \caption{Model Loss}
        \label{fig:modelLoss}
    \end{subfigure}
    \caption{Performance in training model for \texttt{SYNTH} data set.}
    \label{fig:trainingperf}
\end{figure}

The sample code in Figure~\ref{fig:training} creates and trains one of the neural networks. There are five such neural networks, one for each waypoint. First, it finds the number of output neurons needed by finding the number of possible outputs or labels. It incorporates early stopping, which stops the training process when the {\em validation loss} starts to increase. Validation loss describes how far away the model’s predictions are from true labels in a test data set. Early stopping is done to prevent {\em overfitting}, which is when a machine learning model is trained specifically for its training data and performs poorly on unseen data. Next, I performed tests to determine the number of layers and the number of neurons there should be in each layer. The Softmax activation function is used for the output layer and ReLU is used for the rest. For example, for the \texttt{WALKING} data set, the neural net is created with two layers and 300 neurons for the input layer. This was determined after testing values from 50 to 1000 neurons and 2 to 5 layers. 2 layers with an input layer containing 300 neurons was chosen because it was determined to be the least value that results in low percent error in DTW results on 1000 trials as shown in Figure~\ref{fig:neuronError} since the goal is to simplify the model as much as possible to reduce execution time when using the model for prediction. Finally, the model is trained. Figure~\ref{fig:trainingperf} shows the evolution of the model as it is being trained. Figure~\ref{fig:modelAcc} and Figure~\ref{fig:modelLoss} show the accuracy and loss for the training data set and a separate test data set as the model is trained using the \texttt{SYNTH} data set. In this case, we can see that accuracy reaches 0.625 (62.5\%) and loss stabilizes at 1.0. Note that this accuracy is on the specific predictions made for the waypoints. It does not quantify how close the prediction might be if it is wrong. In addition, when the prediction is close, the subsequent steps of the \sys algorithm will still find the optimal warp path.

\begin{figure}[t]
\begin{lstlisting}[language=Python]
mlinput = []
waypoints = ["(0, 0)"]
confidence = [1]

# reformat subsections of time series
mlinput += x1.tolist()[0:length] + x2.tolist()[0:width]
mlinput = np.array(mlinput)        
mlinput = mlinput.reshape(1, -1)

#predict waypoint 0
transformedModelInput = scalerwp0.transform(mlinput)ok
output0 = DTWModel0.predict(transformedModelInput)
output0 = output0.tolist()[0]
waypoints.append(labelMapwp0[output0.index(max(output0))])
confidence.append(max(output0))

# Repeat above with DTWModel[1..4] to predict other 
# waypoints. Code omitted for brevity 

waypoints.append(f"({n-1}, {m-1})")
\end{lstlisting}
\caption{Code predicts waypoints using partial signal measurements.}
\label{fig:mldtwcode1}
\end{figure}

\subsection{Using Predictions}
\label{sec:usingPredictions}

The first step in using the machine learning model is to collect the $n$ points from each time series that is used as input. Next, the code in Figure~\ref{fig:mldtwcode1} reformats the result into a format that can be used by the computed machine learning model. This formatted result (\texttt{mlinput}) is passed to the machine learning model (\texttt{DTWModel0}) to produce a prediction output. This output is appended to the list of known waypoints. Note that the confidence is also saved. The last part of this code (lines 15--19) is repeated with the other models to produce the predictions for the four other waypoints.

\begin{figure}[t]
\begin{lstlisting}[language=Python]
# Compute path of search region
path = [(0, 0)]
for i in range(1, len(waypoints)):
    curRow, curCol = path[-1][0], path[-1][1] 
    wp1 = waypoints[i]
    coords = wp1[1:len(wp1)-1].split(',')
    targetRow = min(max(curRow+1, int(coords[0])), n-1) 
    targetCol = min(max(curCol+1, int(coords[1].strip())), m-1)

    slope = (targetRow - curRow) / (targetCol - curCol)
    lastRow = curRow
    for i in range(targetCol - curCol):
        path.append((lastRow, (i + curCol)))
        if i == targetCol - curCol - 1:
            for j in range(lastRow, targetRow):
                path.append((j, (i + curCol)))
                lastRow = j
        else:
            for j in range(lastRow,round(i*slope)+curRow):
                path.append((j, (i + curCol)))
                lastRow = j
     
# Compute widths of search region
confidenceWidths = [14]
for val in confidence:
    confidenceWidths.append(int(((2-val))*(m/10)))
confidenceWidths.append(14)

widths = []
for i in range(len(confidenceWidths)-1):
    widths.append(confidenceWidths[i])
    slope = (confidenceWidths[i+1]-confidenceWidths[i])/(n//6)
    for j in range(n//6-1):
        widths.append(int(confidenceWidths[i]+slope*j))
widths.append(confidenceWidths[len(confidenceWidths)-1])
widths.append(confidenceWidths[len(confidenceWidths)-1])
\end{lstlisting}
\caption{Code takes the set of waypoint predictions and confidence values to compute a search region.}
\label{fig:mldtwcode2}
\end{figure}

The code in Figure~\ref{fig:mldtwcode2} is used to take the set of waypoint predictions and confidence values to compute a search region. The code is split into two major parts. The first part computes the center of the path of the search region. The code uses the predicted waypoints and identifies the points on the path connecting the waypoints. These points are stored in the variable \texttt{path}. The width of the search path is computed based on the confidence of the machine learning predictions. The start and end width is set to a constant value, 14 in this case. The width of the search area at the waypoints is set proportionally to the confidence of the waypoint. The width at other points in the path is made to linearly adjust the width from one waypoint to the next.

The code for the key step of computing the cost matrix values is shown in Figure~\ref{fig:mldtwcode3}. The first part of this code converts the path and widths computed earlier to an actual range of matrix locations. The matrix values are then computed by the loop at the end. Once the matrix values in the search area are computed, I use the backtracking code shown in Figure~\ref{fig:dtwbackimpl} to compute the warp path.

\begin{figure}[t]
\begin{lstlisting}[language=Python]
for i in range(1, n+1):
    # determine range of cells to compute
    valsInCol = []
    middleVal = 0
    for xVal, yVal in path:
        if xVal+1 == i:
            valsInCol.append(yVal)
    if len(valsInCol) != 0:
        middleVal = sum(valsInCol) // len(valsInCol)
    width2 = widths[i-1]
    if len(valsInCol) >= width2:
        width2 = len(valsInCol) + 1

    if middleVal-(width2//2) < 0:
        xi = 1
        xf = min(m, xi+width2)
    else:
        xf = min(m, middleVal+(width2//2+1))
        xi = xf-width2
    if i == 1:
        xi = 1
        xf = min(m, xi+width2)
    elif i == n:
        xf = m
        xi = xf-width2
        
    # compute dtw matrix values in region
    for j in range(xi, xf+1):
        if dtw_matrix[i, j] == np.inf:
            cost = abs(x1[i-1] - x2[j-1])
            last_min = np.min([dtw_matrix[i-1, j], 
                dtw_matrix[i, j-1], d
                stw_matrix[i-1, j-1]])
            dtw_matrix[i, j] = cost + last_min
            
dist=dtw_matrix[len(dtw_matrix)-1][len(dtw_matrix[0])-1]
\end{lstlisting}
\caption{This code computes the cost matrix values within \sys{}'s constrained search space.}
\label{fig:mldtwcode3}
\end{figure}
\section{Evaluation}
\label{sec:results}

In this section, I explore the performance of \sys using the four data sets described in Section~\ref{sec:dataset} (\texttt{SYNTH, WRITING, ACTIVITY}, and \texttt{WALKING}). Some of the key questions that my evaluation explores include:
\begin{itemize}
    \item Can \sys learn the common properties of warp paths using my proposed approach?
    \item What performance/accuracy tradeoffs does \sys provide? 
    \item How well does \sys work on different real-world applications?
    \item Are there significant differences in \sys{}'s performance across applications?
\end{itemize}

\begin{figure*}[t]
    \begin{subfigure}{0.3\linewidth}
      \centering
      \includegraphics[width=\textwidth]{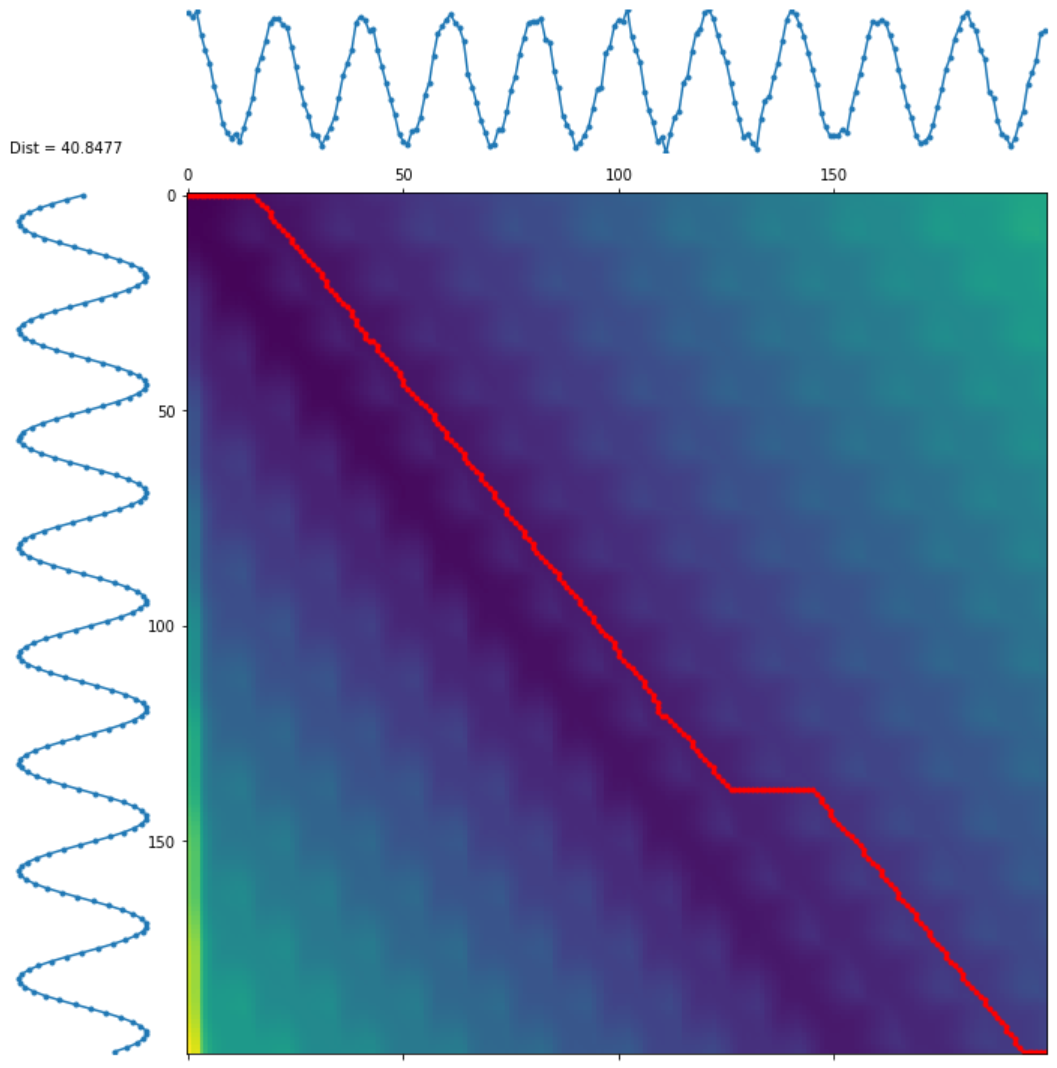} 
      \caption{FullDTW: Distance: 40.8, Time: 0.13s
}
      \label{fig:evalsynthfull}
    \end{subfigure}
    \hfill
    \begin{subfigure}{0.3\linewidth}
      \centering
      \includegraphics[width=\textwidth]{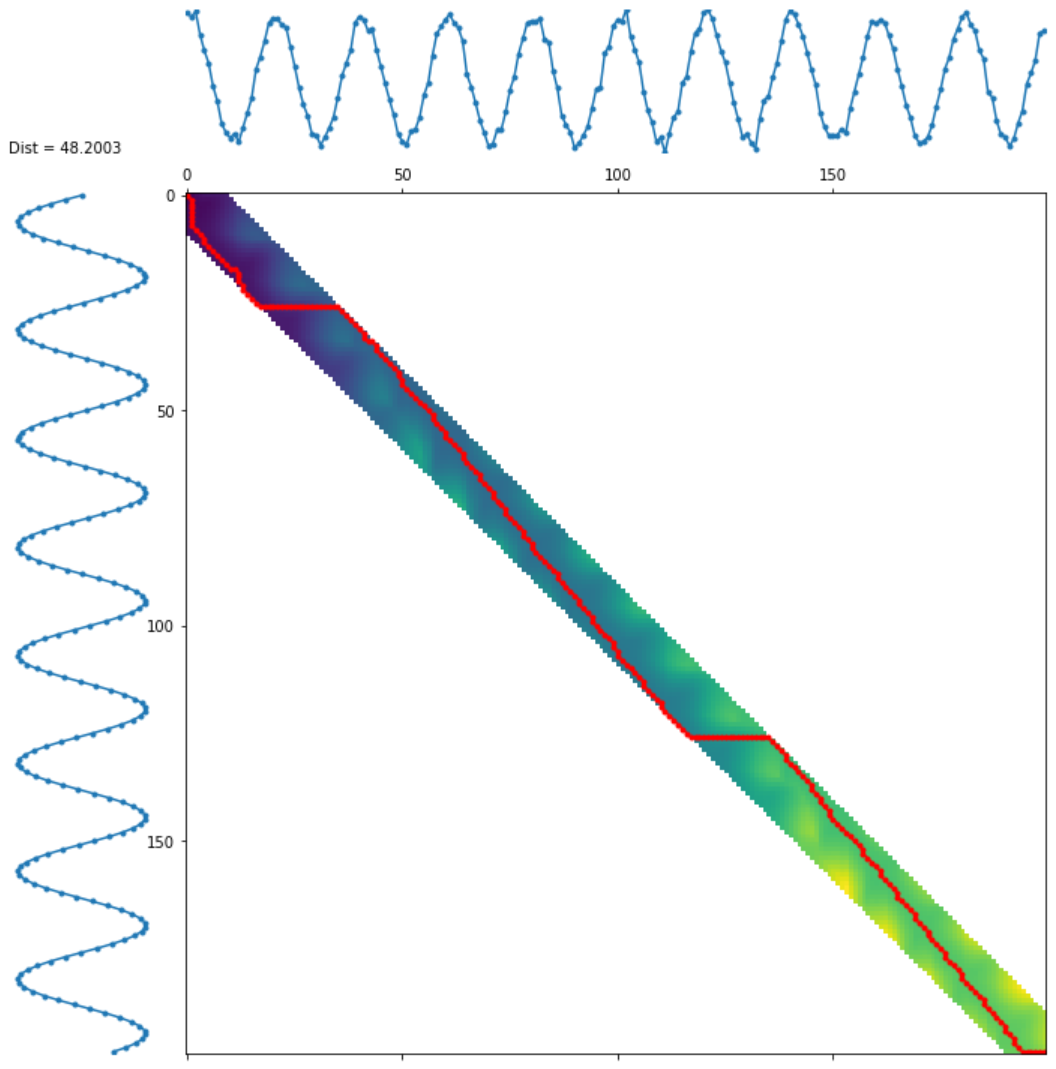} 
      \caption{FastDTW: Distance: 48.2, Error: 18.0\%, Time: 0.016s
}
      \label{fig:evalsynthfast}
    \end{subfigure}
    \hfill
    \begin{subfigure}{0.3\linewidth}
      \centering
      \includegraphics[width=\textwidth]{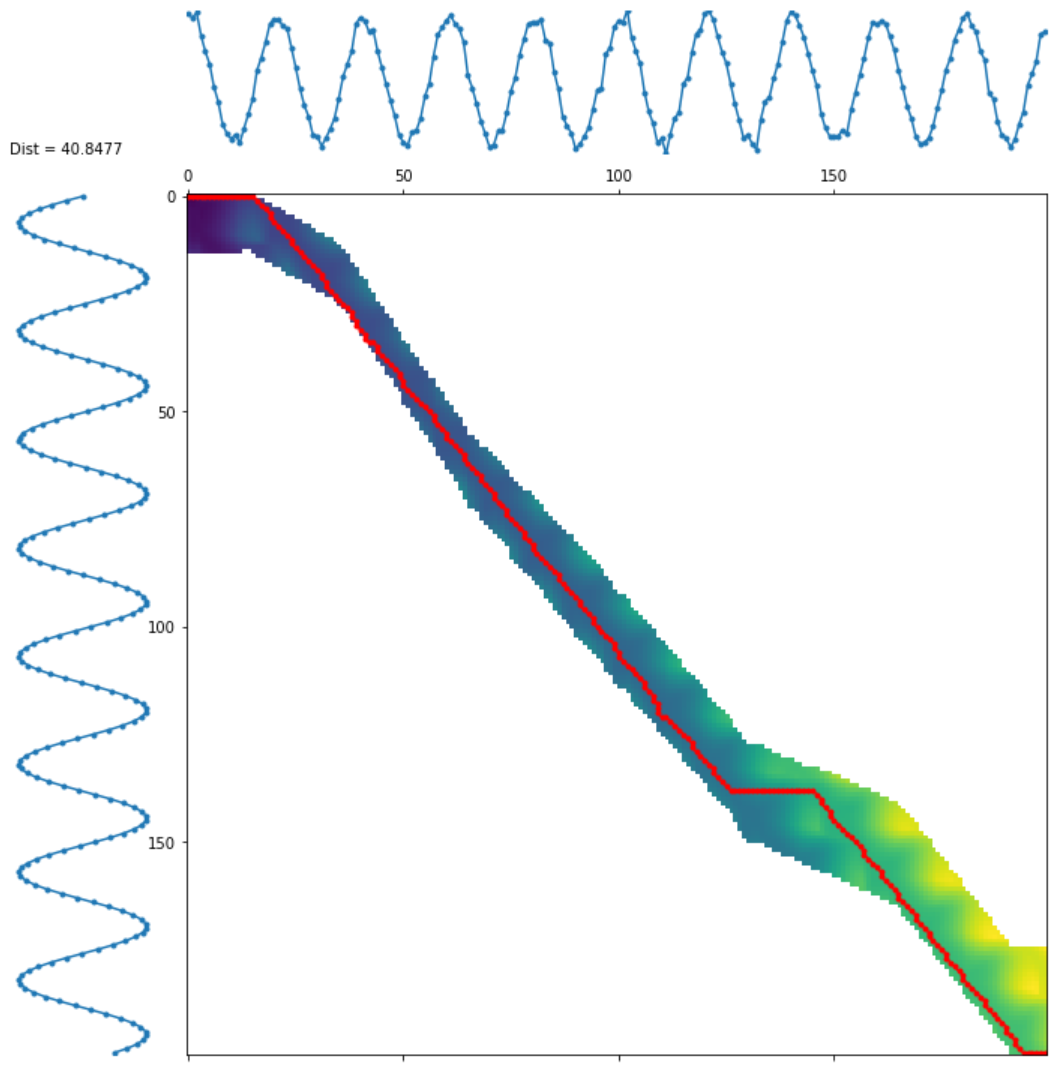} 
      \caption{\sys: Distance: 40.8, Error: 0.0\%, Time: 0.016s, ML Overhead: 0.14s
}
      \label{fig:evalsynthml}
    \end{subfigure}
  \caption{Comparing DTW variants using a single result from the \texttt{SYNTH} data set.}
  \label{fig:evalsynth}
\end{figure*}

\subsection{Synthetic Signals}

I first examine how \sys performs on the \texttt{SYNTH} data set. I also run the FullDTW and FastDTW algorithms on the same signals to compare their accuracy and computation overheads. Recall from Section~\ref{sec:dataset} that the \texttt{SYNTH} exhibits signal noise, shifts, stretches and compression. Figure~\ref{fig:evalsynth} shows a single result from this data set. Figure~\ref{fig:evalsynthfull} shows how FullDTW computes the entire cost matrix. The resulting optimal path has distance 40.8 and follows a path above the diagonal. FastDTW (Figure~\ref{fig:evalsynthfast}) computes only matrix elements down the diagonal. As a result, the calculated distance is larger -- 48.2, an error of 18.0\% from the FullDTW value. In contrast, the result using \sys, shown in Figure~\ref{fig:evalsynthml}, does a good job of predicting that the path is above the diagonal. In addition, it also accurately predicts the slope of the line. In fact, the warp path that is calculated differs only slightly from that calculated by FullDTW. The end result is that \sys has a distance estimate of 40.8 (correctly outputting the true distance) in this example taking the same 0.016 seconds that FastDTW takes. Note there is the additional overhead associated performing inference using the machine learning model. This takes 0.14 seconds in this case. However, this is a constant time overhead regardless of the length of the signals. As a result, the baseline computation times more accurately represent the computational scaling of the algorithms.


I run the same comparison for 1000 pairs of signals from the \texttt{SYNTH} data set. The results of these 1000 trials is that \sys has an median error of 5.88\% from the FullDTW result while FastDTW's error is 13.3\% (Figure~\ref{fig:errors}). In terms of computation time, \sys mas a median time of 0.0163 seconds to compute the necessary distances, while FastDTW takes 0.0164 seconds and FullDTW took 0.129 seconds (Figure~\ref{fig:times}). This is a significant improvement and the gains are potentially much greater on larger signal traces.

The results on the synthetic traces show great promise. \sys is able to learn the types of distortions that are common and identify the distortion in a particular instance using just a small subset of the overall cost matrix. 


\subsection{Real-World Signals}

The above results show that \sys is able to handle the particular range of distortions that I add to the signals in the \texttt{SYNTH} data set. To evaluate if \sys can handle the distortions that may appear in real-world applications, I try applying \sys to two specific application: handwriting recognition and activity recognition. I describe the results from each application below. 

\subsubsection{Activity Recognition}


\begin{figure*}[t]
    \centering
    \begin{subfigure}{0.3\linewidth}
        \includegraphics[width=\textwidth]{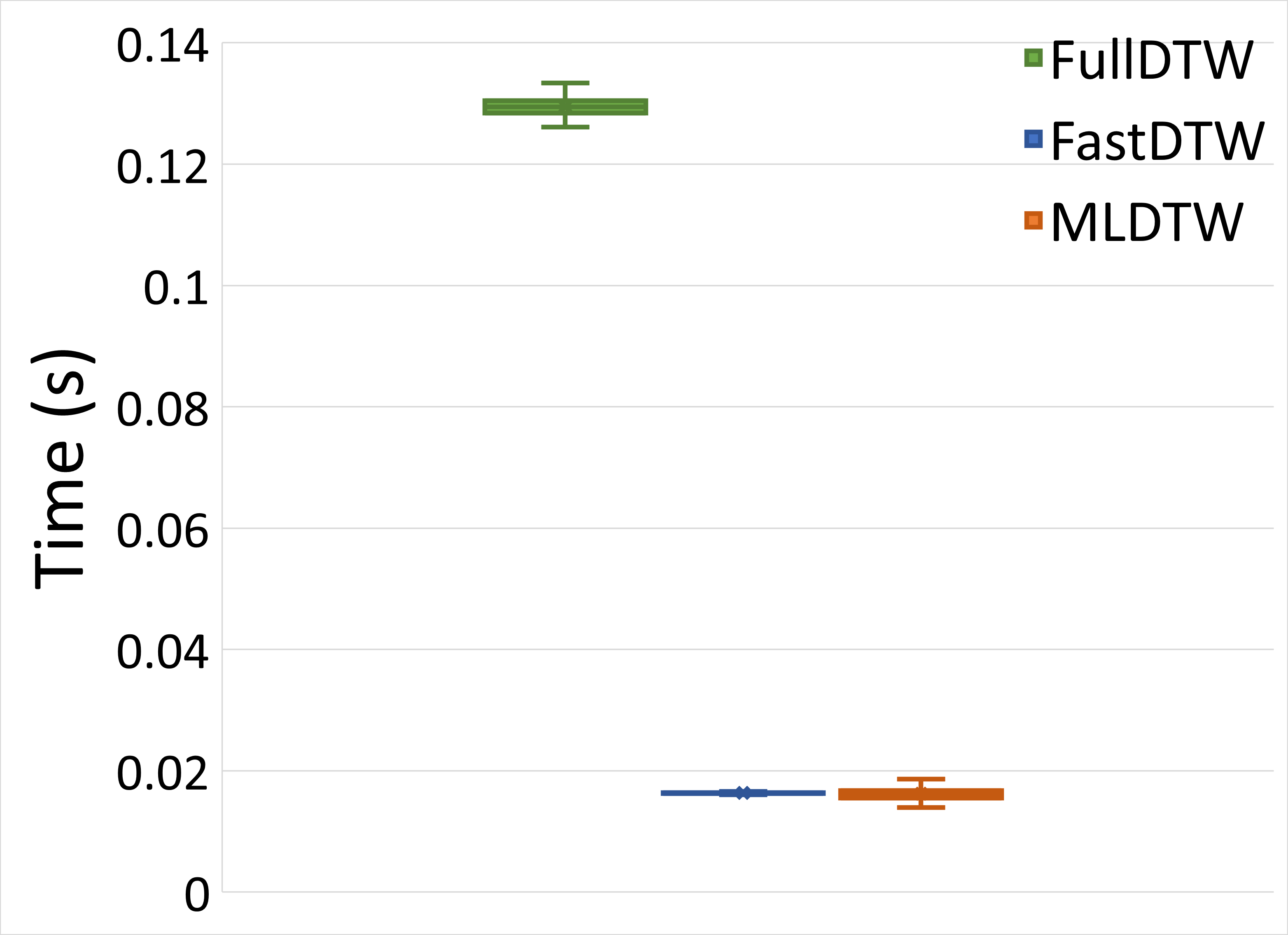} \\  
        \caption{Execution time on \texttt{SYNTH} data set.}
        \label{fig:times}
    \end{subfigure}
    \hfill
    \begin{subfigure}{0.3\linewidth}
        \includegraphics[width=\textwidth]{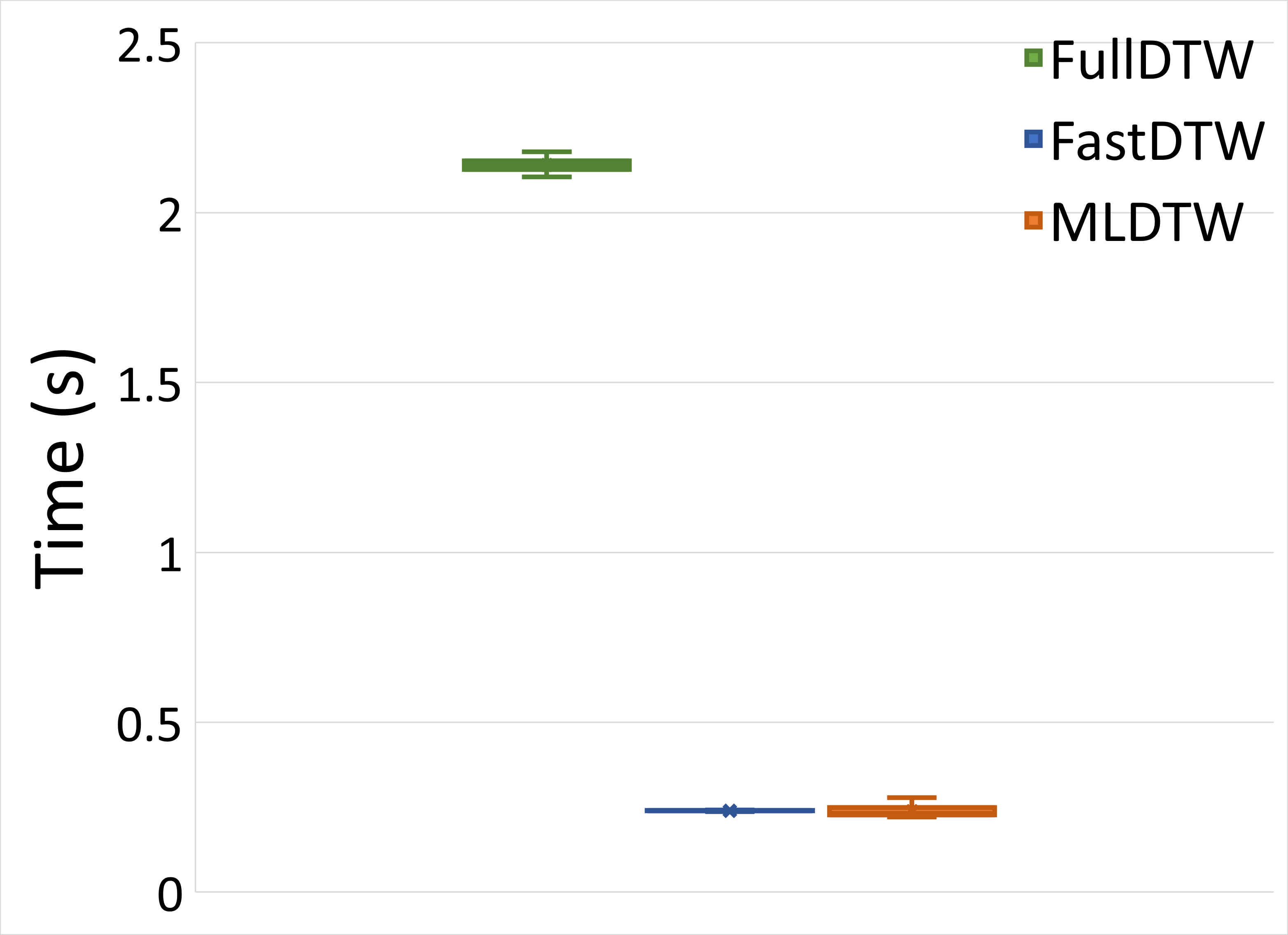} \\  
        \caption{Execution time on \texttt{WALKING} data set.}
        \label{fig:timeacc}
    \end{subfigure}
    \hfill
    \begin{subfigure}{0.3\linewidth}
        \includegraphics[width=\textwidth]{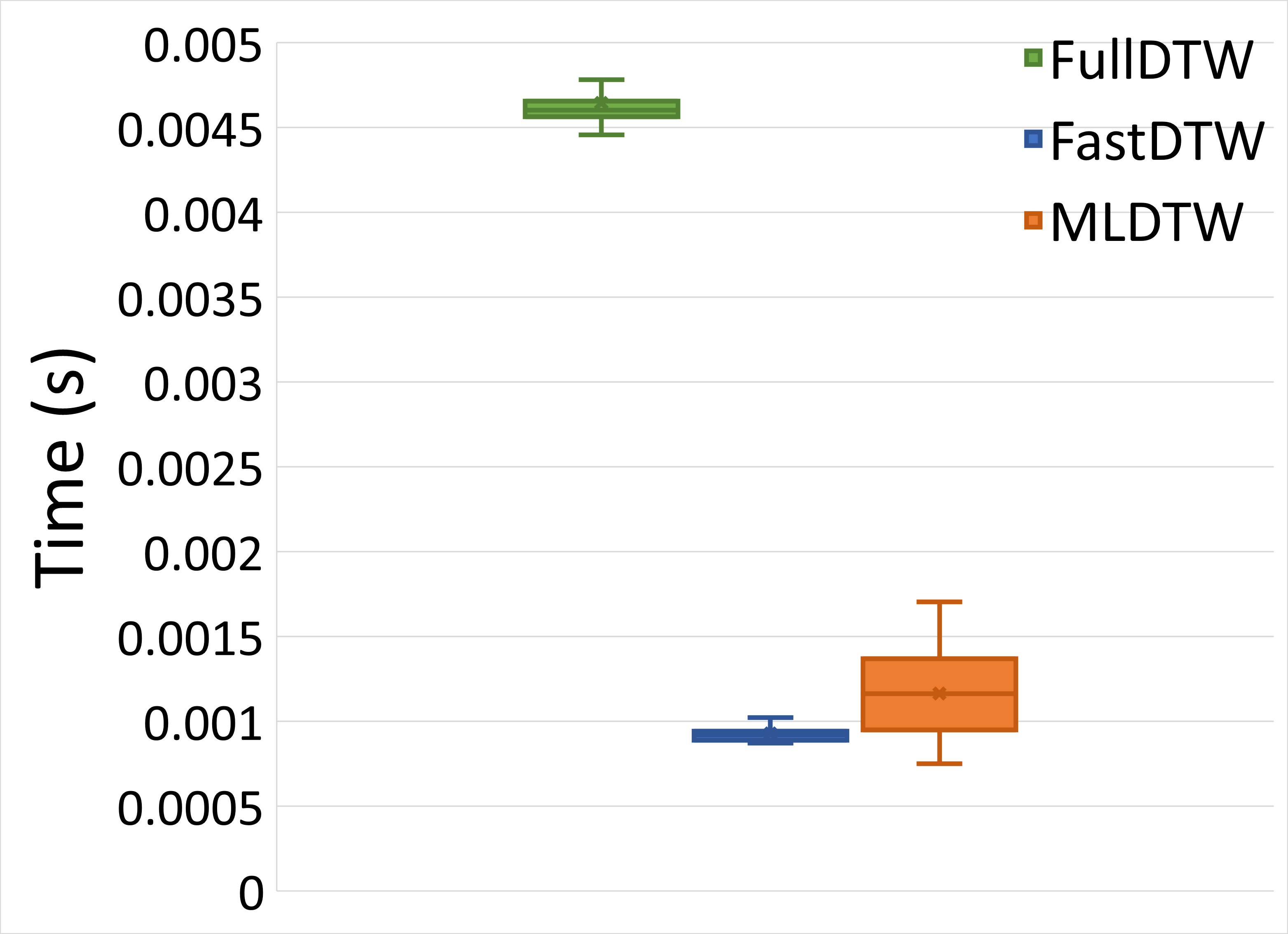} \\  
        \caption{Execution time on \texttt{WRITING} data set.}
        \label{fig:timer}
    \end{subfigure}    
    
    \begin{subfigure}{0.3\linewidth}
        \includegraphics[width=\textwidth]{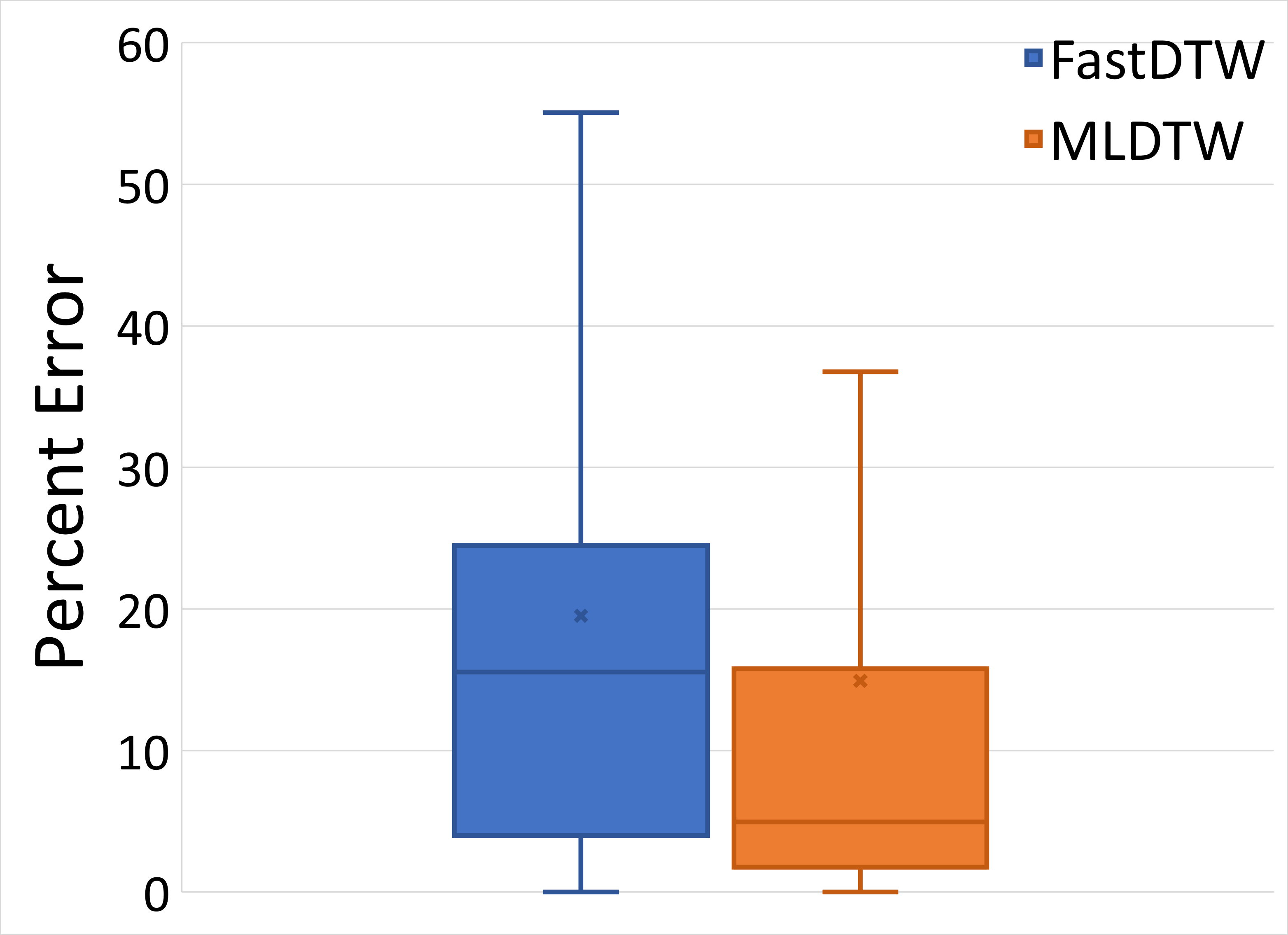} \\  
        \caption{Accuracy on \texttt{SYNTH} data set.}
        \label{fig:errors}
    \end{subfigure}
    \hfill
    \begin{subfigure}{0.3\linewidth}
        \includegraphics[width=\textwidth]{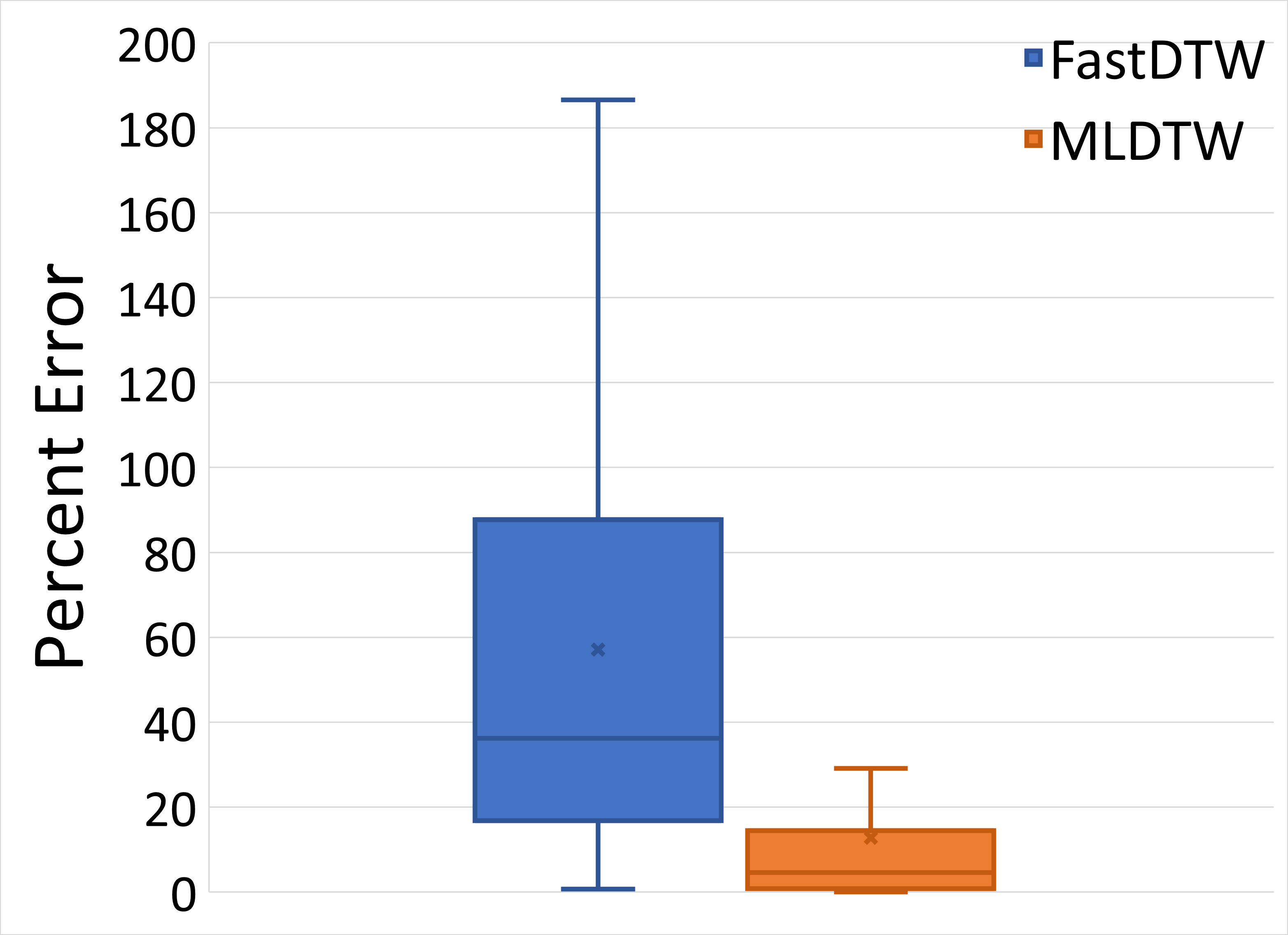} \\  
        \caption{Accuracy on \texttt{WALKING} data set.}
        \label{fig:erroracc}
    \end{subfigure}
    \hfill
    \begin{subfigure}{0.3\linewidth}
        \includegraphics[width=\textwidth]{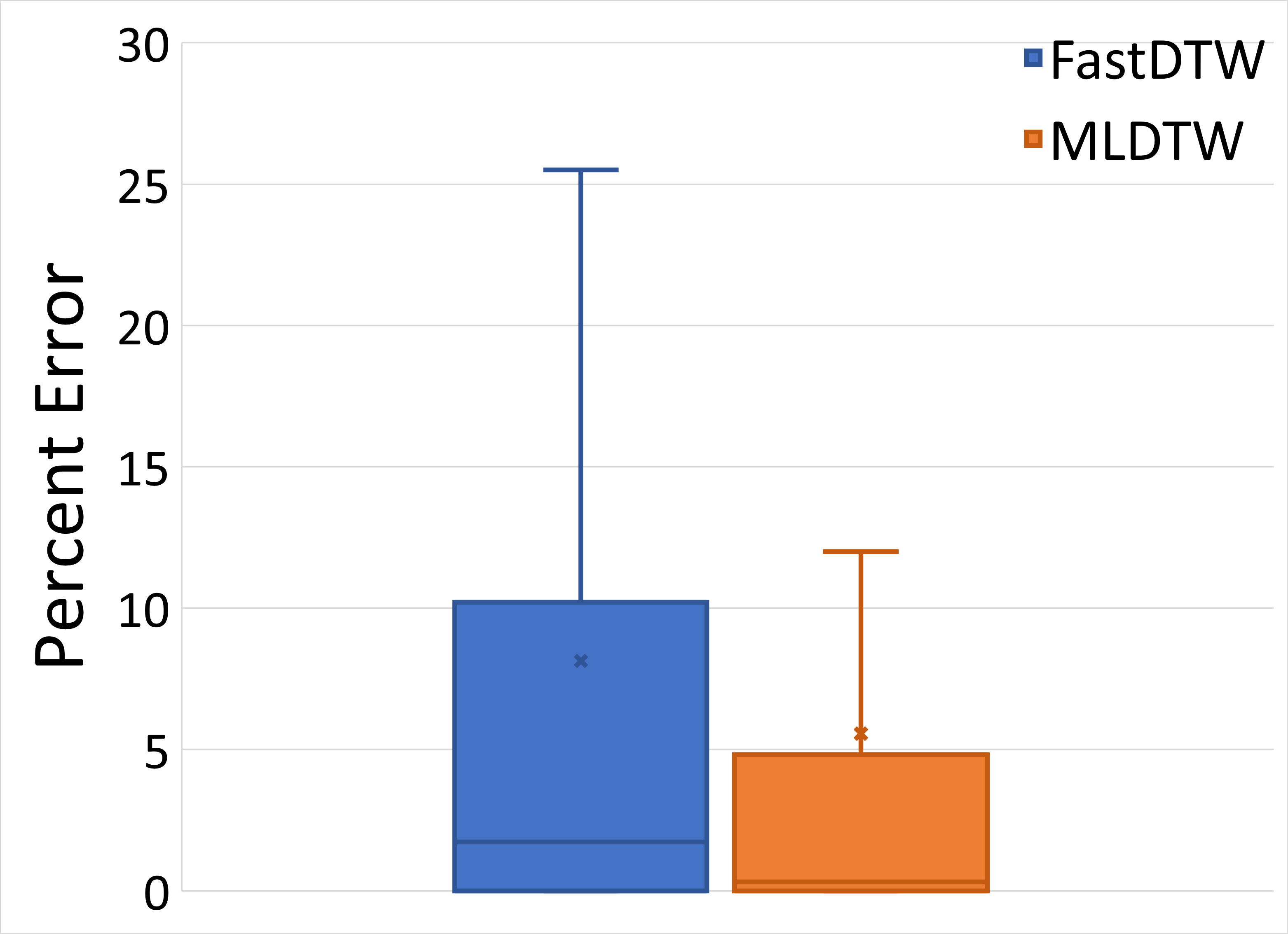} \\  
        \caption{Accuracy on \texttt{WRITING} data set.}
        \label{fig:errorr}
    \end{subfigure}

    \caption{Performance of different DTW variants on \texttt{SYNTH, WALKING,} and \texttt{WRITING} data sets.}
    \label{fig:writing}

\end{figure*}

\begin{figure*}[t]
    \begin{subfigure}{0.3\linewidth}
      \centering
      \includegraphics[width=\textwidth]{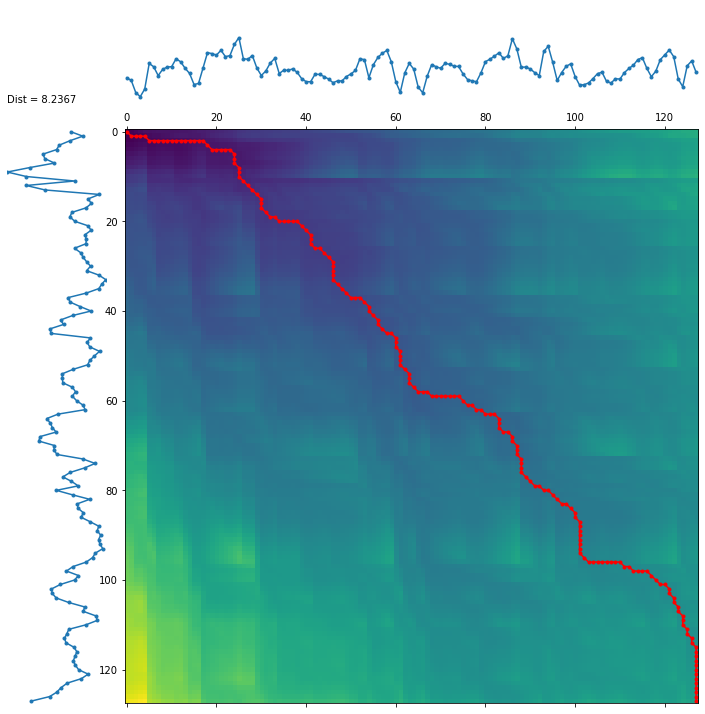} 
      \caption{FullDTW: Distance: 8.24, Time: 0.054s
}
      \label{fig:evalfull2}
    \end{subfigure}
    \hfill
    \begin{subfigure}{0.3\linewidth}
      \centering
      \includegraphics[width=\textwidth]{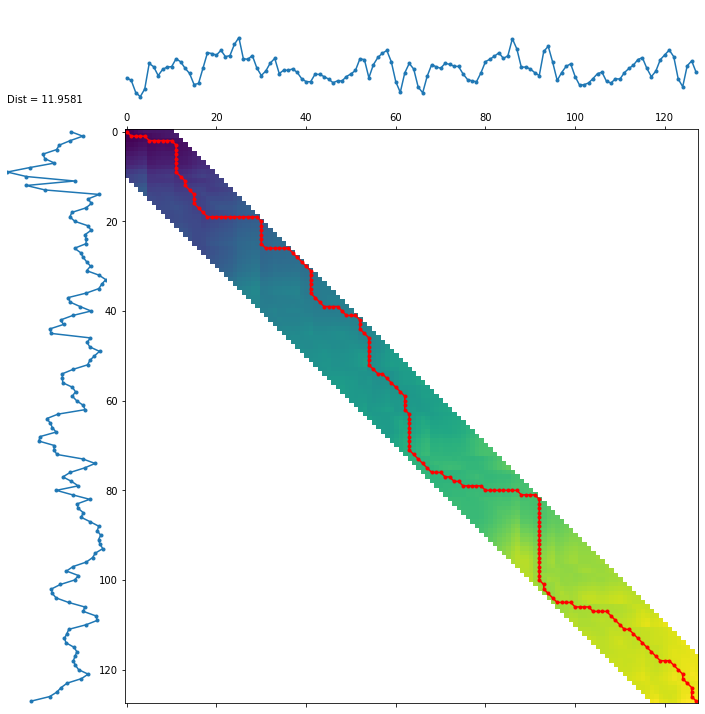} 
      \caption{FastDTW: Distance: 12.0, Error: 45.2\%, Time: 0.011s
}
      \label{fig:evalfast2}
    \end{subfigure}
    \hfill
    \begin{subfigure}{0.3\linewidth}
      \centering
      \includegraphics[width=\textwidth]{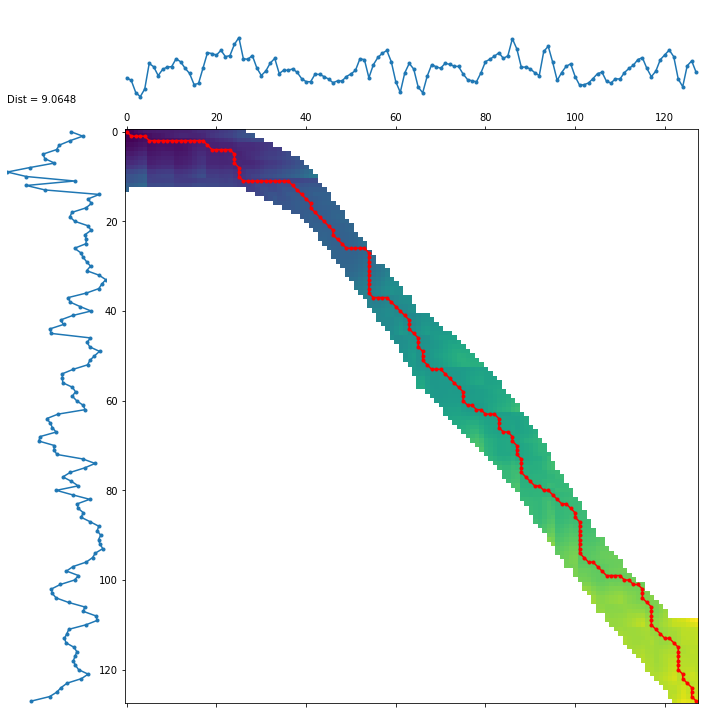} 
      \caption{\sys: Distance: 9.06, Error: 10.1\%, Time: 0.086s, ML Overhead: 0.14s }
      \label{fig:evalml2}
    \end{subfigure}
  \caption{Comparing DTW variants using a single result from the \texttt{ACTIVITY} data set
}
  \label{fig:eval2}
\end{figure*}

I use two data sets for activity recognition testing -- \texttt{ACTIVITY} and \texttt{WALKING}.

Figure~\ref{fig:eval2} illustrates the results from using the various DTW algorithms on a specific pair of acceleration measurements from the \texttt{ACTIVITY} data set. As seen in Figure~\ref{fig:evalfull2}, FullDTW computes the optimal warp distance of 8.24 and takes a time of 0.054 seconds. Note that the time is much lower than FullDTW on the \texttt{SYNTH} data set, where it took 0.13 seconds to compute. The reason is that the signals in \texttt{ACTIVITY} are only 128 samples long while the signals in \texttt{SYNTH} were 200 samples long. This is a difference of a factor of 1.56. However, since FullDTW has computational complexity of $\mathcal{O}(n^2)$, this means that it will take $1.56^2 = 2.44$ times as long. This closely predicts the time difference since $\frac{0.13}{0.054} = 2.41$. The signal variation over time is also far less predictable than the sine waves used in \texttt{SYNTH}. This results in less structure to the cost matrix and a slightly more unpredictable shape to the optimal warp path. Despite this, \sys is able to roughly predict the correct shape in Figure~\ref{fig:evalml2} and have distance estimate that is only off by 10.1\%. Compared to Figure~\ref{fig:evalsynth}, \sys{}'s use of confidence to adjust the size of the search region is much more visible in this example. Also, in contrast to \sys, FastDTW's use of the diagonal results in a poor estimate of the warp distance with an error of 45.2\%.

I ran one hundred trials similar to the one shown in Figure~\ref{fig:eval2}. Across these trials, \sys had a lower average error and time than FastDTW. \sys has 19.9\% error in 0.009 seconds compared to 23.7\% error in 0.011 seconds for FastDTW and 0.054 seconds for FullDTW. This shows that \sys provides a significant improvement over current techniques even on real-world data sets. 




The \texttt{WALKING} data set consists of accelerometer readings from two different individuals, both walking with the accelerometer in their pocket. To eliminate phone orientation as a factor, the absolute magnitude of acceleration is used. Each signal measurement consists of 4 seconds of accelerometer readings taken at 200Hz. Figures~\ref{fig:timeacc} and \ref{fig:erroracc} summarize the computation time and accuracy, respectively, of comparing 100 different pairs of measurements. \sys had a median accuracy of 4.60\% error, which is much lower than FastDTW's 36.2\% error. \sys had a median time of 0.23 seconds compared to FastDTW's 0.24 seconds per comparison while FullDTW took around 2.1 seconds.

\subsubsection{Handwriting Recognition}





The \texttt{WRITING} data set provides a second application from a very different domain. This provides an opportunity to see if \sys is able to identify common warp patterns of different types. For example, while the differences in accelerometer readings in activity recognition likely stem from the physical differences between users, the differences in writing likely stem from differences in how users were trained to draw particular letters. Sample results for recognizing the letter "a" using the data in the \texttt{WRITING} data set are presented here. Figures~\ref{fig:timer} and \ref{fig:errorr} show the computation time and accuracy of the different DTW algorithms on this comparison. \sys provides lower error than FastDTW: 0.32\% vs. 1.72\% as shown in Figure~\ref{fig:errorr}. The computation time used \sys is 0.0011 seconds per comparison rather than 0.0046 and 0.00093 taken by FastDTW and FullDTW, respectively. This shows that \sys is able to identify and make use of the predictability in warp patterns for drawing letters to improve accuracy. However, it is notable that MLDTW is slightly slower than FastDTW in this case. In general, some of the fixed overheads in MLDTW result in longer execution times on smaller signal traces. While the magnitude of the accuracy gain is small, MLDTW is able to eliminate many of the outlier values that would result in misidentification of the handwriting. This is also reflected in the significant relative reduction in error. 

\section{Discussion}
\label{sec:discussion}

In this section, I discuss some of the limitations of \sys and identify potential areas for future improvement.

{\em Machine Learing Overhead.} One drawback of \sys is that the overhead associated with performing inference to predict the waypoints can be significant. There are two aspects to this overhead: the number of inputs to the model and the complexity of the model.

First, in \sys, the input to the machine learning model must contain a sufficient portion of each signal to be able to identify key factors that impact the warp path. For example, in an application such as activity recognition where stride length plays a critical role, the subsection of each time series may need to contain a full stride. It will be valuable to explore the relationship between the length of the subsections, accuracy, and computation time.

Second, the machine learning model can be relatively complex and take time to perform inference using the model. To account for this, \sys could leverage history. More specifically, for a specific user, the warps are likely consistent between different invocations of \sys. For example, for a user who generally talks slowly, DTW will always warp in a particular way to compensate for the slow speech. \sys could incorporate a mechanism to identify these user-specific warp paths and use FastDTW once they are identified. This hybrid approach may perform better by providing the accuracy benefits of \sys without the associated inference overhead. Finally, there is growing availability of inference accelerators on many types of user devices. \sys could easily make use of such hardware to execute inference more quickly. 

\paragraph{Application Domains.} In this paper, I explored two real-world application domains: activity recognition and handwriting recognition. \sys works best for applications where there are common factors that impact the optimal warp path. In the case of activity recognition and handwriting recognition, physical differences between users and different writing styles are such factors. One important area of future work is to identify further applications where \sys works well.

\begin{figure}[t]
    \begin{subfigure}{0.9\linewidth}
      \centering
      \includegraphics[width=\textwidth]{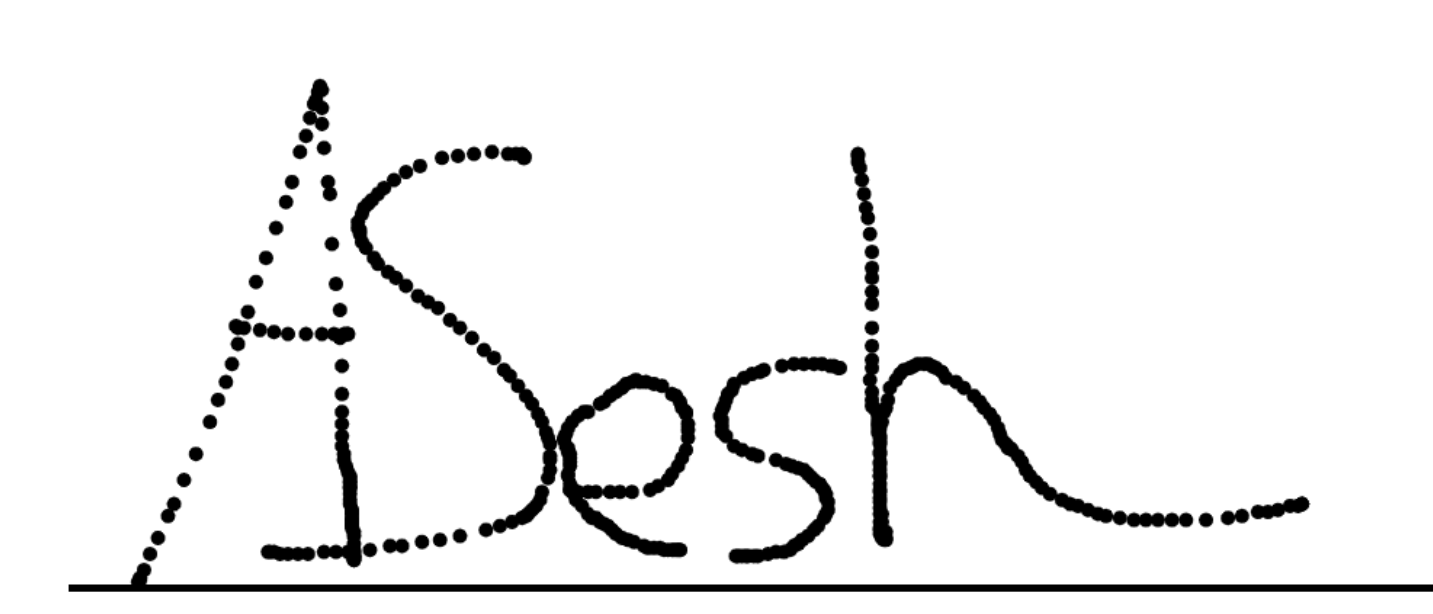} 
      \caption{Application provides a canvas where user can draw their signature.}
      \label{fig:sig}
    \end{subfigure}
    \begin{subfigure}{0.9\linewidth}
        \centering
        \includegraphics[width=\textwidth]{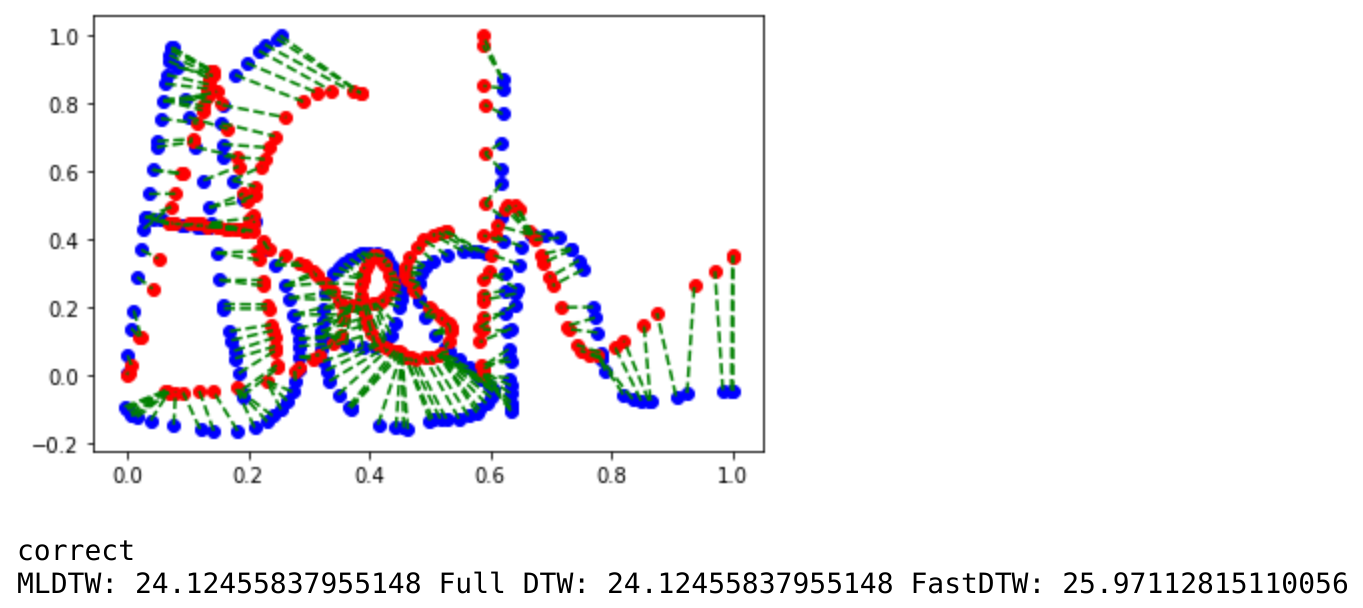} \\  
        \caption{DTW algorithms compute the match between drawn signatures and pre-existing samples. The application outputs "correct", "maybe correct" or "incorrect" based on the computed DTW distance.}
        \label{fig:dtwsig}
    \end{subfigure}
  \caption{Signature Recognition Application}
  \label{fig:sigapp}
\end{figure}

\paragraph{Application Integration.} In this paper, I focused on the low-level performance of different DTW algorithms on different data sets. To observe the effective use of these algorithms in real-world applications, I have built a prototype application that performs signature recognition using MLDTW. A screenshot of the operation of this application is shown in Figure~\ref{fig:sigapp}. A key next step is to incorporate MLDTW into other applications.

\section{Conclusion}
\label{sec:conclusion}

The results of my study indicate that \sys can have valuable benefits over FastDTW in performing dynamic time warping by significantly improving the accuracy while maintaining or reducing the computational overhead of the computation. The most obvious impact that \sys has is that many applications that require some way to compare signals or already use some variant of dynamic time warping can immediately take advantage of the accuracy and performance benefits the \sys provides. This means that applications such as activity recognition or handwriting recognition will often just work better. A longer term and potentially more significant impact of \sys is to enable new types of applications and qualitative changes in existing application design. \sys{}'s benefits are most significant on applications that need accuracy and also process signals with many values (e.g., due to high sampling frequency). As a result, the improvements of MLDTW can have a significant impact on a variety of applications such as speech recognition or authentication, energy consumption analysis, and health monitoring. Its scalability benefits can also enable new applications such as faster multivariate DTW and higher frequency and more accurate sensing.

%

\bibliographystyle{ACM-Reference-Format}
\bibliography{_references}

\newpage
\onecolumn
\appendix

\section{Implementation}
\label{sec:code}

\subsection{Labeling Data for Training}
The code below corresponds to the tasks described in Section~\ref{sec:labels}. Specifically, it generates a set of five waypoints along the optimal warp path and the $n \times n$ cost matrix as indicated in Figure~\ref{fig:mltrain}.

\label{sec:collectioncode}
\begin{lstlisting}[language=Python]
import math
from dtaidistance import dtw
from dtaidistance import dtw_visualisation as dtwvis
import numpy as np
import csv
import os
import random

def myround(x, base):
    return base * round(x/base)

label = ""

activity  = "walking"

length = 30
width = 30
length -= 1
width -= 1

file = f'acc_{activity}.csv'
if(os.path.exists(file) and os.path.isfile(file)):
    os.remove(file)

header = ""
for i in range(length):
    for j in range(width):
        header += f"col {i}: {j}, "
for i in range(5):
    header += f"waypoints {i}, "
header = header.split(", ")
header.pop()
file = open(f'acc_{activity}.csv','w', newline='')
with file:
    writer = csv.writer(file)
    writer.writerow(header)

for c1 in range(len(trainingData)):
    for c2 in range(len(trainingData)):
        if c1 == c2:
            continue
        x1 = np.asarray(trainingData[c1])
        x2 = np.asarray(trainingData[c2])

        n = len(x1)
        m = len(x2)

        dtw_matrix = np.zeros((n+1, m+1))
        for i in range(n+1):
            for j in range(m+1):
                dtw_matrix[i, j] = np.inf
        dtw_matrix[0, 0] = 0
        for i in range(1, n+1):
            for j in range(1, m+1):
                cost = abs(x1[i-1] - x2[j-1])
                last_min = np.min([dtw_matrix[i-1, j], dtw_matrix[i, j-1], dtw_matrix[i-1, j-1]])
                dtw_matrix[i, j] = cost + last_min
        dist = dtw_matrix[len(dtw_matrix)-1][len(dtw_matrix[0])-1]
        best_path = dtw.best_path(dtw_matrix)

        dtw_matrix_2 = []
        dtw_matrix = dtw_matrix[0:length+1]
        for row in dtw_matrix:
            dtw_matrix_2.append(row[0:width+1])
        dtw_matrix_2 = np.asarray(dtw_matrix_2)


        line = []
        line += x1.tolist()[0:length] + x2.tolist()[0:width]
        waypoints = []
        for i in range(0, m, m//6):
            vals = []
            for row, col in best_path:
                if col == i:
                    vals.append(row)
            waypoints.append((sum(vals)//len(vals), i))

        waypoints.pop(0)
        waypoints.pop()
        for wprow, wpcol in waypoints:
            line.append((myround(wprow, 5), myround(wpcol, 5)))
        if len(waypoints) == 5:
            file = open(f'acc_{activity}.csv','a', newline='')
            with file:
                writer = csv.writer(file)
                writer.writerow(line)
\end{lstlisting}

\subsection{Model Training}
\label{sec:trainingcode}

The below code shows a complete implementation of model training described in Section~\ref{sec:training}.

\begin{lstlisting}[language=Python]

from tensorflow.keras import models
from tensorflow.keras import layers
from tensorflow.keras.callbacks import EarlyStopping

outputNeurons0 = len(np.unique(y0))

early_stopping0 = EarlyStopping(monitor='val_loss', patience=10)

model0 = models.Sequential()
model0.add(layers.Dense(500,  activation='relu', input_shape =(X_train_ML_0.shape[1],)))
model0.add(layers.Dense(outputNeurons0,  activation='softmax'))
model0.compile(optimizer='adam',
              loss='sparse_categorical_crossentropy',
              metrics =['accuracy'])
history0 = model0.fit(X_train_ML_0,
                    y_train_ML_0,
                    epochs=200,
                    validation_data =(X_test_ML_0, y_test_ML_0),
                    callbacks=[early_stopping0])
model0.save(f'acc_{activity}_0.h5')

# Repeat above with DTWModel [1..4] to train models for the other four waypoints. Code omitted for brevity
\end{lstlisting}

\subsection{Feature Prediction}
\label{sec:extractioncode}
The below code predicts waypoints and uses these predictions to generate a warp path. These steps are described in Section~\ref{sec:usingPredictions}.

\begin{lstlisting}[language=Python]
dtw_matrix = np.zeros((n+1, m+1))
for i in range(n+1):
    for j in range(m+1):
        dtw_matrix[i, j] = np.inf
dtw_matrix[0, 0] = 0

for i in range(1, length+1):
    for j in range(1, width+1):
        cost = abs(x1[i-1] - x2[j-1])
        last_min = np.min([dtw_matrix[i-1, j], dtw_matrix[i, j-1], dtw_matrix[i-1, j-1]])
        dtw_matrix[i, j] = cost + last_min
dist = dtw_matrix[len(dtw_matrix)-1][len(dtw_matrix[0])-1]

trainingFlip = dtw_matrix.transpose()
mlinput = []
for i in range(length):
    column = trainingFlip[i+1]
    column = column[1:width+1]
    for val in column:
        mlinput.append(val)
confidence = []
mlinput = np.array(mlinput)        
mlinput = mlinput.reshape(1, -1)
waypoints = []
waypoints.append("(0, 0)")
confidence.append(1)

transformedModelInput = scalerwp0.transform(mlinput)
output0 = DTWModel0.predict(transformedModelInput)
output0 = output0.tolist()[0]
waypoints.append(labelMapwp0[output0.index(max(output0))])
confidence.append(max(output0))

# Repeat above with DTWModel [1..4] to predict other waypoints. Code omitted for brevity

waypoints.append(f"({n-1}, {m-1})")
path = [(0, 0)]
waypointCols = []
for i in range(1, len(waypoints)):
    wp0 = waypoints[i-1]
    curRow, curCol = path[-1][0], path[-1][1] 
    waypointCols.append(curCol)
    wp1 = waypoints[i]
    coords = wp1[1:len(wp1)-1].split(',')
    targetRow, targetCol = min(max(curRow+1, int(coords[0])), n-1), min(max(curCol+1, int(coords[1].strip())), m-1)
    if (targetCol - curCol) <= 0:
        while curRow != targetRow:
            path.append((curRow, curCol))
            curRow += 1
    else:
        slope = (targetRow - curRow) / (targetCol - curCol)
        slope = max(0, slope)
        lastRow = curRow
        for i in range(targetCol - curCol):
            path.append((lastRow, (i + curCol)))
            if i == targetCol - curCol - 1:
                for j in range(lastRow, targetRow):
                    path.append((j, (i + curCol)))
                    lastRow = j
            else:
                for j in range(lastRow, round(i*slope)+curRow):
                    path.append((j, (i + curCol)))
                    lastRow = j
                
    best_path = []
    for i in path:
        if i not in best_path:
            best_path.append(i)
confidenceWidths = []
widths = []
for val in confidence:
    confidenceWidths.append(int(((2-val))*(m/10)))
confidenceWidths.append(14)
confidenceWidths.insert(0, 14)
for i in range(len(confidenceWidths)-1):
    widths.append(confidenceWidths[i])
    slope = (confidenceWidths[i+1]-confidenceWidths[i])/(n//6)
    for j in range(n//6-1):
        widths.append(int(confidenceWidths[i]+slope*j))
widths.append(confidenceWidths[len(confidenceWidths)-1])
widths.append(confidenceWidths[len(confidenceWidths)-1])
subStartTime = time.time()
for i in range(1, n+1):
    valsInCol = []
    middleVal = 0
    for xVal, yVal in best_path:
        if xVal+1 == i:
            valsInCol.append(yVal)
    if len(valsInCol) != 0:
        middleVal = sum(valsInCol) // len(valsInCol)
    width2 = widths[i-1]
    if len(valsInCol) >= width2:
        width2 = len(valsInCol) + 1

    if middleVal-(width2//2) < 0:
        xi = 1
        xf = min(m, xi+width2)
    else:
        xf = min(m, middleVal+(width2//2+1))
        xi = xf-width2
    if i == 1:
        xi = 1
        xf = min(m, xi+width2)
    elif i == n:
        xf = m
        xi = xf-width2
    for j in range(xi, xf+1):
        if dtw_matrix[i, j] == np.inf:
            cost = abs(x1[i-1] - x2[j-1])
            last_min = np.min([dtw_matrix[i-1, j], dtw_matrix[i, j-1], dtw_matrix[i-1, j-1]])
            dtw_matrix[i, j] = cost + last_min
dist1 = dtw_matrix[len(dtw_matrix)-1][len(dtw_matrix[0])-1]
\end{lstlisting}

\subsection{Acceleration Collection Android Application}
\label{sec:acccollectionapp}

The below code is used on an Android phone to collect data for the \texttt{WALKING} dataset. This dataset is described in Section~\ref{sec:dataset}.

\begin{lstlisting}[language=Java]

package com.example.readsensors;

import androidx.appcompat.app.AppCompatActivity;
import android.content.Context;
import android.hardware.SensorEventListener;
import android.hardware.SensorManager;
import android.os.Bundle;
import android.hardware.Sensor;
import android.hardware.SensorEvent;
import android.os.Environment;
import android.view.View;
import android.widget.TextView;
import java.io.File;
import java.io.FileOutputStream;
import java.io.IOException;


public class MainActivity extends AppCompatActivity implements SensorEventListener {
    private SensorManager senSensorManager;
    private Sensor senAccelerometer;
    public static float x;
    public static float y;
    public static float z;
    public static boolean record = false;
    public static String data = "time, x, y, z" + "\r\n";


    @Override
    protected void onCreate(Bundle savedInstanceState) {
        super.onCreate(savedInstanceState);
        setContentView(R.layout.activity_main);

        senSensorManager = (SensorManager) getSystemService(Context.SENSOR_SERVICE);
        senAccelerometer = senSensorManager.getDefaultSensor(Sensor.TYPE_ACCELEROMETER);
        senSensorManager.registerListener(this, senSensorManager.getDefaultSensor(Sensor.TYPE_ACCELEROMETER), 5000);

    }

    Context context;

    @Override
    public void onSensorChanged(SensorEvent event) {
        Sensor mySensor = event.sensor;

        if (mySensor.getType() == Sensor.TYPE_ACCELEROMETER) {
            x = event.values[0];
            y = event.values[1];
            z = event.values[2];
        }

        if (record) {
            data = data + String.valueOf(System.currentTimeMillis()) + "," + new String(String.valueOf(x)) + "," + new String(String.valueOf(y)) + "," + new String(String.valueOf(z)) + "\n";
        }
    }

    @Override
    public void onAccuracyChanged(Sensor sensor, int accuracy) {

    }

    protected void onPause() {
        super.onPause();

    }

    protected void onResume() {
        super.onResume();

    }

    public void startRecording(View view) {
        record = true;
        TextView textViewMessage = findViewById(R.id.textViewMessage);
        textViewMessage.setText("recording");
    }

    public void stopRecording(View view) {
        record = false;
        data = data + "\r\n";
        TextView textViewMessage = findViewById(R.id.textViewMessage);
        textViewMessage.setText("waiting");
    }

    public void sendData(View view) throws IOException {
        String FILENAME = "phone_data" + System.currentTimeMillis() / 1000L + ".csv";

        File folder = Environment.getExternalStoragePublicDirectory(Environment.DIRECTORY_DOWNLOADS);
        File myFile = new File(folder, FILENAME);
        FileOutputStream fstream = new FileOutputStream(myFile);
        fstream.write(data.getBytes());
        fstream.close();
        TextView textViewMessage = findViewById(R.id.textViewMessage);
        textViewMessage.setText("worked");
        data = "time, x, y, z" + "\r\n";
    }
}
\end{lstlisting}

\subsection{Other Code}
I have omitted code for the following tasks for space:
\begin{itemize}
    \item FastDTW Implementation
    \item Synthetic Dataset Generation
    \item Data Extraction From Text Files
    \item Training and Inference for Regression Approach
    \item Signature Recognition Application
    \item \texttt{WRITING} Data Set Collection Application
\end{itemize}

\end{document}